\documentclass{article}

\PassOptionsToPackage{numbers, sort}{natbib}
 \usepackage[final]{nips_2016}

\usepackage[utf8]{inputenc} 
\usepackage[T1]{fontenc}    
\usepackage{hyperref}       
\usepackage{url}            
\usepackage{booktabs}       
\usepackage{amsfonts}       
\usepackage{nicefrac}       
\usepackage{microtype}      

\usepackage{times}
\usepackage{graphicx} 
\usepackage{subfigure} 

\usepackage{amsmath}
\usepackage{amsfonts}
\usepackage{amssymb, amsthm, bm}
\newtheorem{theorem}{Theorem}
\newtheorem{prop}{Proposition}

\newtheorem{corollary}{Corollary}

\DeclareMathOperator*{\argmax}{arg\,max}
\usepackage{algorithm}
\usepackage{algorithmic}

\renewcommand{\arraystretch}{0.94}
\usepackage{enumitem}

\usepackage{capt-of}
\usepackage{pdfpages}

\title{R{\'e}nyi Divergence Variational Inference}

\author{
  Yingzhen Li \\
  University of Cambridge\\
  Cambridge, CB2 1PZ, UK \\
  \texttt{yl494@cam.ac.uk} \\
  \And
  Richard E.~Turner \\
  University of Cambridge\\
  Cambridge, CB2 1PZ, UK \\
  \texttt{ret26@cam.ac.uk} \\
}

\begin{document}

\maketitle

\begin{abstract}
This paper introduces the \emph{variational R{\'e}nyi bound (VR)} that extends traditional variational inference to R{\'e}nyi's $\alpha$-divergences. This new family of variational methods unifies a number of existing approaches, and enables a smooth interpolation from the evidence lower-bound to the log (marginal) likelihood that is controlled by the value of $\alpha$ that parametrises the divergence. The reparameterization trick, Monte Carlo approximation and stochastic optimisation methods are deployed to obtain a tractable and unified framework for optimisation. We further consider negative $\alpha$ values and propose a novel variational inference method as a new special case in the proposed framework. Experiments on Bayesian neural networks and variational auto-encoders demonstrate the wide applicability of the VR bound.
\end{abstract}

\section{Introduction}
Approximate inference, that is approximating posterior distributions and likelihood functions, is at the core of modern probabilistic machine learning. This paper focuses on optimisation-based approximate inference algorithms, popular examples of which include variational inference (VI), variational Bayes (VB) \cite{jordan:vi, beal:thesis} and expectation propagation (EP) \cite{minka:ep, opper:ec}. Historically, VI has received more attention compared to other approaches, although EP can be interpreted as iteratively minimising a set of local divergences \cite{minka:divergence}. This is mainly because VI has elegant and useful theoretical properties such as the fact that it proposes a lower-bound of the log-model evidence. Such a lower-bound can serve as a surrogate to both maximum likelihood estimation (MLE) of the hyper-parameters and posterior approximation by Kullback-Leibler (KL) divergence minimisation.


Recent advances of approximate inference follow three major trends.
First, scalable methods, e.g.~stochastic variational inference (SVI) \cite{hoffman:svi} and stochastic expectation propagation (SEP) \cite{li:sep, barthelme:aep}, have been developed for datasets comprising millions of datapoints. Recent approaches \cite{broderick:stream, gelman:dep, xu:sms} have also applied variational methods to coordinate parallel updates arising from computations performed on chunks of data.
Second, Monte Carlo methods and black-box inference techniques have been deployed to assist variational methods, e.g.~see \cite{paisley:bbvi, salimans:reparam, ranganath:bbvi, kucukelbir:vi_stan} for VI and \cite{hernandez-lobato:bb-alpha} for EP. They all proposed ascending the Monte Carlo approximated variational bounds to the log-likelihood using noisy gradients computed with automatic differentiation tools.
Third, tighter variational lower-bounds have been proposed for (approximate) MLE. The importance weighted auto-encoder (IWAE) \cite{burda:iwae} improved upon the variational auto-encoder (VAE) \cite{kingma:vae, rezende:vae} framework, by providing tighter lower-bound approximations to the log-likelihood using importance sampling. 
These recent developments are rather separated and little work has been done to understand their connections.

In this paper we try to provide a unified framework from an energy function perspective that encompasses a number of recent advances in variational methods, and we hope our effort could potentially motivate new algorithms in the future. This is done by extending traditional VI to R{\'e}nyi's $\alpha$-divergence \cite{renyi:divergence}, a rich family that includes many well-known divergences as special cases. After reviewing useful properties of R{\'e}nyi divergences and the VI framework, we make the following contributions:

\begin{itemize}[leftmargin=8mm]
 \item We introduce the \emph{variational R{\'e}nyi bound} (VR) as an extension of VI/VB. We then discuss connections to existing approaches, including VI/VB, VAE, IWAE \cite{burda:iwae}, SEP \cite{li:sep} and black-box alpha (BB-$\alpha$) \cite{hernandez-lobato:bb-alpha}, thereby showing the richness of this new family of variational methods.
 \item We develop an optimisation framework for the VR bound. An analysis of the bias introduced by stochastic approximation is also provided with theoretical guarantees and empirical results.
 \item We propose a novel approximate inference algorithm called \emph{VR-max} as a new special case. Evaluations on VAEs and Bayesian neural networks show that this new method is often comparable to, or even better than, a number of the state-of-the-art variational methods.
\end{itemize}

\section{Background}
This section reviews R{\'e}nyi's $\alpha$-divergence and variational inference upon which the new framework is based. Note that there exist other $\alpha$-divergence definitions \citep{amari:divergence, tsallis:divergence} (see appendix). However we mainly focus on R{\'e}nyi's definition as it enables us to derive a new class of variational lower-bounds.

\subsection{R{\'e}nyi's $\alpha$-divergence}

\label{sec:renyi_divergence}
We first review R{\'e}nyi's $\alpha$-divergence \cite{renyi:divergence, van_erven:renyi}. R{\'e}nyi's $\alpha$-divergence, defined on $\{\alpha: \alpha > 0, \alpha \neq 1, |\mathrm{D}_{\alpha}| < +\infty \}$, measures the ``closeness'' of two distributions $p$ and $q$ on a random variable $\bm{\theta} \in \Theta$:
\begin{equation}
\label{eq:renyi_divergence}
\mathrm{D}_{\alpha} [p || q] = \frac{1}{\alpha - 1} \log \int p(\bm{\theta})^{\alpha} q(\bm{\theta})^{1 - \alpha} d \bm{\theta}.
\end{equation}
The definition is extended to $\alpha = 0, 1, +\infty$ by continuity. We note that when $\alpha \rightarrow 1$ the Kullback-Leibler (KL) divergence is recovered, which plays a crucial role in machine learning and information theory. Some other special cases are presented in Table \ref{tab:renyi_example}. The method proposed in this work also considers $\alpha \leq 0$ (although (\ref{eq:renyi_divergence}) is no longer a divergence for these $\alpha$ values), and we include from \cite{van_erven:renyi} some useful properties for forthcoming derivations.
\begin{prop}
(Monotonicity) R{\'e}nyi's $\alpha$-divergence definition (\ref{eq:renyi_divergence}), extended to negative $\alpha$, is \textbf{continuous} and \textbf{non-decreasing} on $\alpha \in \{\alpha: -\infty < \mathrm{D}_{\alpha} < +\infty \}$.
\label{prop:renyi_divergence}
\end{prop}
\begin{prop}
(Skew symmetry) For $\alpha \not\in \{0, 1\}$, 
$
\mathrm{D}_{\alpha} [p || q] = \frac{\alpha}{1 - \alpha} \mathrm{D}_{1 - \alpha} [q || p].
$
This implies $\mathrm{D}_{\alpha} [p || q] \leq 0$ for $\alpha < 0$. For the limiting case $\mathrm{D}_{-\infty} [p || q] = -\mathrm{D}_{+\infty} [q || p]$.
\label{prop:skew_symmetry}
\end{prop}
\begin{table}[t]
  \centering
  \caption{Special cases in the R{\'e}nyi divergence family.}
  \renewcommand{\arraystretch}{1.2}
  \label{tab:renyi_example}
  \begin{tabular}{ccl}
  	\toprule
    $\alpha$ & Definition & Notes\\
    \hline
    $\alpha \rightarrow 1$ & $\int p(\bm{\theta}) \log \frac{p(\bm{\theta})}{q(\bm{\theta})} d \bm{\theta}$ & 
    \begin{tabular}{@{}l@{}} \emph{Kullback-Leibler (KL) divergence}, \\ used in VI ($\mathrm{KL}[q||p]$) and EP ($\mathrm{KL}[p||q]$) \end{tabular} \\
    $\alpha = 0.5$ & $-2 \log (1 - \mathrm{Hel}^2[p||q])$ & function of the square \emph{Hellinger distance}\\
    $\alpha \rightarrow 0$ & $-\log \int_{p(\bm{\theta}) > 0} q(\bm{\theta}) d\bm{\theta}$ & 
    \begin{tabular}{@{}l@{}} zero when $\mathrm{supp}(q) \subseteq \mathrm{supp}(p)$ \\ (not a divergence) \end{tabular} \\
    $\alpha = 2$ & $-\log (1 - \chi^2[p||q])$ & 
    \begin{tabular}{@{}l@{}} proportional to the $\chi^2$-divergence \end{tabular} \\
    $\alpha \rightarrow +\infty$ & $\log \max_{\bm{\theta} \in \Theta} \frac{p(\bm{\theta})}{q(\bm{\theta})}$ &
    \begin{tabular}{@{}l@{}} \emph{worst-case regret} in \\ \emph{minimum description length principle} \cite{grunwald:mdl}\end{tabular} \\
    \bottomrule
  \end{tabular}
\end{table}
A critical question that is still in active research is how to choose a divergence in this rich family to obtain optimal solution for a particular application, an issue which is discussed in the appendix.

\subsection{Variational inference}
\label{sec:vi}
Next we review the variational inference algorithm \cite{jordan:vi, beal:thesis} using posterior approximation as a running example.
Consider observing a dataset of $N$ i.i.d.~samples $\mathcal{D} = \{\bm{x}_n \}_{n=1}^N$ from a probabilistic model $p(\bm{x}|\bm{\theta})$ parametrised by a random variable $\bm{\theta}$ that is drawn from a prior $p_0(\bm{\theta})$. Bayesian inference involves computing the posterior distribution of the parameters given the data, 
\begin{equation}
p(\bm{\theta} | \mathcal{D}, \bm{\varphi}) = \frac{p(\bm{\theta}, \mathcal{D}| \bm{\varphi})}{p(\mathcal{D}| \bm{\varphi})} = \frac{p_0(\bm{\theta}| \bm{\varphi}) \prod_{n=1}^{N} p(\bm{x}_n | \bm{\theta}, \bm{\varphi})}{p(\mathcal{D}| \bm{\varphi})}, 
\end{equation}
where $p(\mathcal{D}| \bm{\varphi}) = \int p_0(\bm{\theta}| \bm{\varphi}) \prod_{n=1}^{N} p(\bm{x}_n | \bm{\theta}, \bm{\varphi}) d\bm{\theta}$ is called marginal likelihood or model evidence. The hyper-parameters of the model are denoted as $\bm{\varphi}$ which might be omitted henceforth for notational ease. 
For many powerful models the exact posterior is typically intractable, and approximate inference introduces an approximation $q(\bm{\theta})$ in some tractable distribution family $\mathcal{Q}$ to the exact posterior. One way to obtain this approximation is to minimise the KL divergence $\mathrm{KL}[q(\bm{\theta}) || p(\bm{\theta} | \mathcal{D})]$, which is also intractable due the difficult term $p(\mathcal{D})$. Variational inference (VI) sidesteps this difficulty by considering an equivalent optimisation problem that maximises the \emph{variational lower-bound}:
\begin{equation}
\mathcal{L}_{\text{VI}}(q; \mathcal{D}, \bm{\varphi}) = \log p(\mathcal{D}|\bm{\varphi}) - \mathrm{KL}[q(\bm{\theta}) || p(\bm{\theta} | \mathcal{D}, \bm{\varphi})]
	= \mathbb{E}_{q} \left[ \log  \frac{p(\bm{\theta}, \mathcal{D}|\bm{\varphi})}{q(\bm{\theta})} \right].
\label{eq:vi_bound}
\end{equation}
The variational lower-bound can also be used to optimise the hyper-parameters $\bm{\varphi}$.

\begin{figure}[t]
 \centering
 \subfigure[Approximated posterior.]{
 \includegraphics[width=0.24\linewidth]{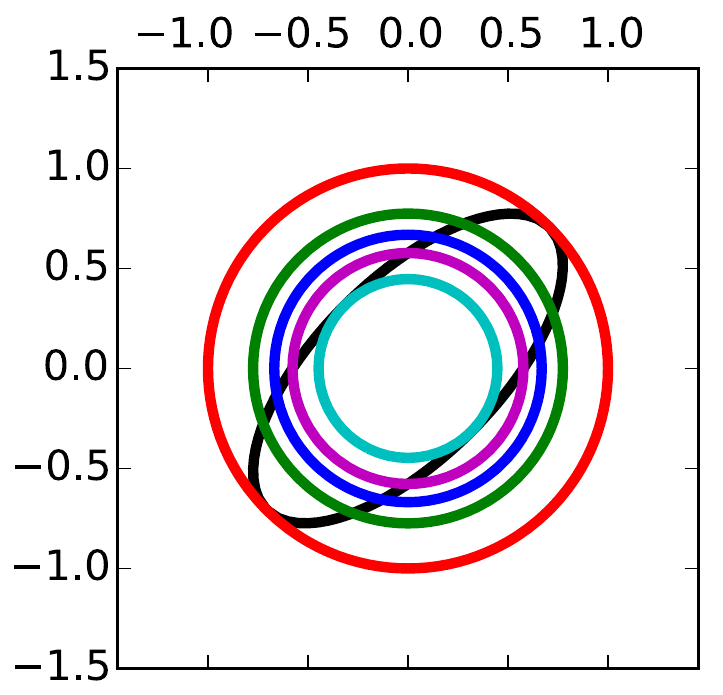}
 \label{fig:linear_regression_posterior}}
 \hspace{0.5in}
 \subfigure[Hyper-parameter optimisation.]{
 \includegraphics[width=0.45\linewidth]{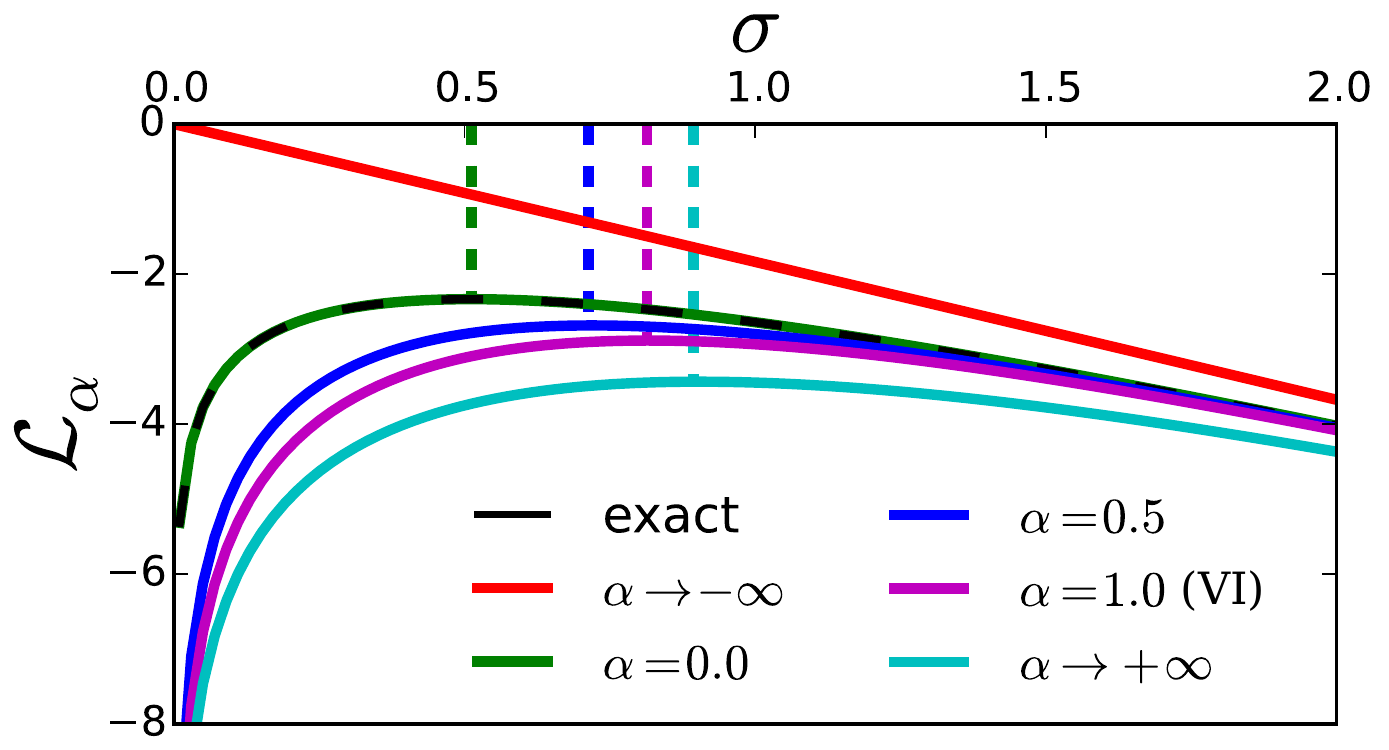}
 \label{fig:linear_regression_energy}}
 \caption{Mean-Field approximation for Bayesian linear regression. In this case $\bm{\varphi} = \sigma$ the observation noise variance. The bound is tight as $\sigma \rightarrow +\infty$, biasing the VI solution to large $\sigma$ values.}
\end{figure}

To illustrate the approximation quality of VI we present a mean-field approximation example to Bayesian linear regression in Figure \ref{fig:linear_regression_posterior} (in magenta). Readers are referred to the appendix for details, but essentially a factorised Gaussian approximation is fitted to the true posterior, a correlated Gaussian in this case. The approximation recovers the posterior mean correctly, but is over-confident. Moreover, as $\mathcal{L}_{\text{VI}}$ is the difference between the marginal likelihood and the KL divergence, hyper-parameter optimisation can be biased away from the exact MLE towards the region of parameter space where the KL term is small \cite{turner:two_problems} (see Figure \ref{fig:linear_regression_energy}). 

\section{Variational R{\'e}nyi bound}
\label{sec:vr_bound}
Recall from Section \ref{sec:renyi_divergence} that the family of R{\'e}nyi divergences includes the KL divergence. Perhaps variational free-energy approaches can be generalised to the R{\'e}nyi case? Consider approximating the exact posterior $p(\bm{\theta}|\mathcal{D})$ by minimizing R{\'e}nyi's $\alpha$-divergence $\mathrm{D}_{\alpha}[q(\bm{\theta}) || p(\bm{\theta} | \mathcal{D})]$ for some selected $\alpha > 0$.
Now we consider the equivalent optimization problem $\max_{q \in \mathcal{Q}} \log p(\mathcal{D}) - \mathrm{D}_{\alpha}[q(\bm{\theta}) || p(\bm{\theta} | \mathcal{D})]$, and when $\alpha \neq 1$, whose objective can be rewritten as
\begin{equation}
\mathcal{L}_{\alpha}(q; \mathcal{D}) := \frac{1}{1 - \alpha} \log \mathbb{E}_{q} \left[ \left( \frac{p(\bm{\theta}, \mathcal{D})}{q(\bm{\theta})} \right)^{1 - \alpha} \right].
\label{eq:alpha_vi}
\end{equation}
We name this new objective the \emph{variational R{\'e}nyi (VR) bound}. Importantly the above definition can be extend to $\alpha \leq 0$, and the following theorem is a direct result of Proposition \ref{prop:renyi_divergence}.
\begin{theorem}
\label{thm:alpha_vi}
The objective $\mathcal{L}_{\alpha}(q; \mathcal{D})$ is \textbf{continuous} and \textbf{non-increasing} on $\alpha \in \{\alpha: |\mathcal{L}_{\alpha}| < +\infty \}$. Especially for all $0 < \alpha_{+} < 1$ and $\alpha_{-} < 0$,
\begin{equation*}
\mathcal{L}_{\text{VI}}(q; \mathcal{D}) = \lim_{\alpha \rightarrow 1} \mathcal{L}_{\alpha}(q; \mathcal{D}) 
 \leq \mathcal{L}_{\alpha_{+}}(q; \mathcal{D}) \leq \mathcal{L}_{0}(q; \mathcal{D}) \leq \mathcal{L}_{\alpha_{-}}(q; \mathcal{D})
\end{equation*}
Also $\mathcal{L}_{0}(q; \mathcal{D}) = \log p(\mathcal{D})$ if and only if the support $ \mathrm{supp}(p(\bm{\theta}|\mathcal{D})) \subseteq  \mathrm{supp}(q(\bm{\theta})) $.
\end{theorem}
Theorem \ref{thm:alpha_vi} indicates that the VR bound can be useful for model selection by sandwiching the marginal likelihood with bounds computed using positive and negative $\alpha$ values, which we leave to future work. In particular $\mathcal{L}_{0} = \log p(\mathcal{D})$ under the mild assumption that $q$ is supported where the exact posterior is supported. This assumption holds for many commonly used distributions, e.g.~Gaussians are supported on the entire space, and in the following we assume that this condition is satisfied. 

Choosing different alpha values allows the approximation to balance between zero-forcing ($\alpha \rightarrow +\infty$, when using uni-modal approximations it is usually called mode-seeking) and mass-covering ($\alpha \rightarrow -\infty$) behaviour. This is illustrated by the Bayesian linear regression example, again in Figure \ref{fig:linear_regression_posterior}. First notice that $\alpha \rightarrow +\infty$ (in cyan) returns non-zero uncertainty estimates (although it is more over-confident than VI) which is different from the maximum a posteriori (MAP) method that only returns a point estimate. Second, setting $\alpha = 0.0$ (in green) returns $q(\bm{\theta}) = \prod_i p(\theta_i|\mathcal{D})$ and the exact marginal likelihood $\log p(\mathcal{D})$ (Figure \ref{fig:linear_regression_energy}). Also the approximate MLE is less biased for $\alpha = 0.5$ (in blue) since now the tightness of the bound is less hyper-parameter dependent.

\section{The VR bound optimisation framework}
This section addresses several issues of the VR bound optimisation by proposing further approximations. First when $\alpha \neq 1$, the VR bound is usually just as intractable as the marginal likelihood for many useful models. However Monte Carlo (MC) approximation is applied here to extend the set of models that can be handled. The resulting method can be applied to any model that MC-VI \cite{paisley:bbvi, salimans:reparam, ranganath:bbvi, kucukelbir:vi_stan} is applied to.
Second, Theorem \ref{thm:alpha_vi} suggests that the VR bound is to be minimised when $\alpha < 0$, which performs disastrously in MLE context. As we shall see, this issue is solved also by the MC approximation under certain conditions. Third, a mini-batch training method is developed for large-scale datasets in the posterior approximation context. Hence the proposed optimisation framework of the VR bound enables tractable application to the same class of models as SVI.

\subsection{Monte Carlo approximation of the VR bound}
\label{sec:sampling}
Consider learning a latent variable model with MLE as a running example, where the model is specified by a conditional distribution $p(\bm{x}|\bm{h}, \bm{\varphi})$ and a prior $p(\bm{h}| \bm{\varphi})$ on the latent variables $\bm{h}$. Examples include models treated by the variational auto-encoder (VAE) approach \cite{kingma:vae, rezende:vae} that parametrises the likelihood with a (deep) neural network. MLE requires $\log p(\bm{x})$ which is obtained by marginalising out $\bm{h}$ and is often intractable, so the VR bound is considered as an alternative optimisation objective. However instead of using exact bounds, a simple Monte Carlo (MC) method is deployed, which uses finite samples $\bm{h}_k \sim q(\bm{h}|\bm{x}), k = 1, ..., K$ to approximate $\mathcal{L}_{\alpha} \approx \hat{\mathcal{L}}_{\alpha, K}$:
\begin{equation}
\label{eq:sampling_estimate}
\hat{\mathcal{L}}_{\alpha, K}(q; \bm{x}) = \frac{1}{1 - \alpha} \log \frac{1}{K} \sum_{k=1}^K \left[\left( \frac{p(\bm{h}_k, \bm{x})}{q(\bm{h}_k|\bm{x})} \right)^{1 - \alpha} \right].
\end{equation}
The importance weighted auto-encoder (IWAE) \cite{burda:iwae} is a special case of this framework with $\alpha = 0$ and $K < +\infty$. But unlike traditional VI, here the MC approximation is biased. Fortunately we can characterise the bias by the following theorems proved in the appendix.
\begin{theorem}
\label{thm:sampling_bound}
Assume $\mathbb{E}_{\{\bm{h}_k\}_{k=1}^K} [ |\hat{\mathcal{L}}_{\alpha, K}(q; \bm{x})| ] < + \infty$ and $|\mathcal{L}_{\alpha}| < +\infty$. Then $\mathbb{E}_{\{\bm{h}_k\}_{k=1}^K} [ \hat{\mathcal{L}}_{\alpha, K}(q; \bm{x}) ]$ as a function of $\alpha \in \mathbb{R}$ and $K \geq 1$ is: \\
1) \textbf{non-decreasing} in $K$ for fixed $\alpha \leq 1$, and \textbf{non-increasing} in $K$ for fixed $\alpha \geq 1$; \\
2) $\mathbb{E}_{\{\bm{h}_k\}_{k=1}^K} [ \hat{\mathcal{L}}_{\alpha, K}(q; \bm{x}) ] \rightarrow \mathcal{L}_{\alpha}$ as $K \rightarrow +\infty$; \\
3) \textbf{continuous} and \textbf{non-increasing} in $\alpha$ with fixed $K$.
\end{theorem}
\begin{corollary}
\label{thm:alpha_k_existence}
For finite $K$, either $\mathbb{E}_{\{\bm{h}_k\}_{k=1}^K} [ \hat{\mathcal{L}}_{\alpha, K}(q; \bm{x}) ] < \log p(\bm{x})$ for all $\alpha$, or there exists $\alpha_K \leq 0$ such that $\mathbb{E}_{\{\bm{h}_k\}_{k=1}^K} [ \hat{\mathcal{L}}_{\alpha_K, K}(q; \bm{x}) ] = \log p(\bm{x})$ and $\mathbb{E}_{\{\bm{h}_k\}_{k=1}^K} [ \hat{\mathcal{L}}_{\alpha, K}(q; \bm{x}) ] > \log p(\bm{x})$ for all $\alpha < \alpha_K$. Also $\alpha_K$ is \textbf{non-decreasing} in $K$ if exists, with $\lim_{K \rightarrow 1} \alpha_K = -\infty$ and $\lim_{K \rightarrow +\infty} \alpha_K = 0$.
\end{corollary}

\begin{figure*}[t]
 \centering
 \subfigure[MC approximated VR bounds.]{
 \includegraphics[width=0.6\linewidth]{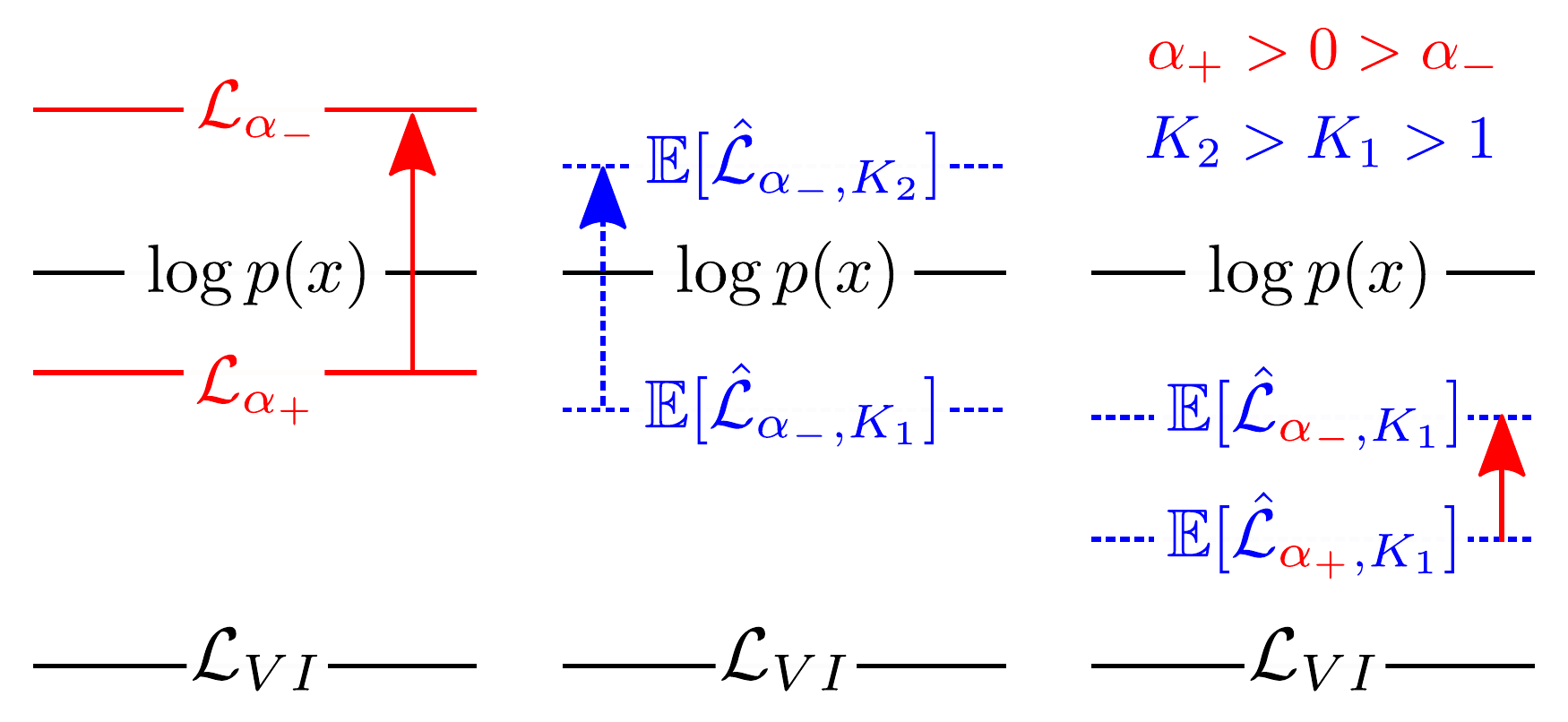}
 \label{fig:divergence_sampling}}
 \hspace{0.1in}
 \subfigure[Simulated MC approximations.]{
 \includegraphics[width=0.34\linewidth]{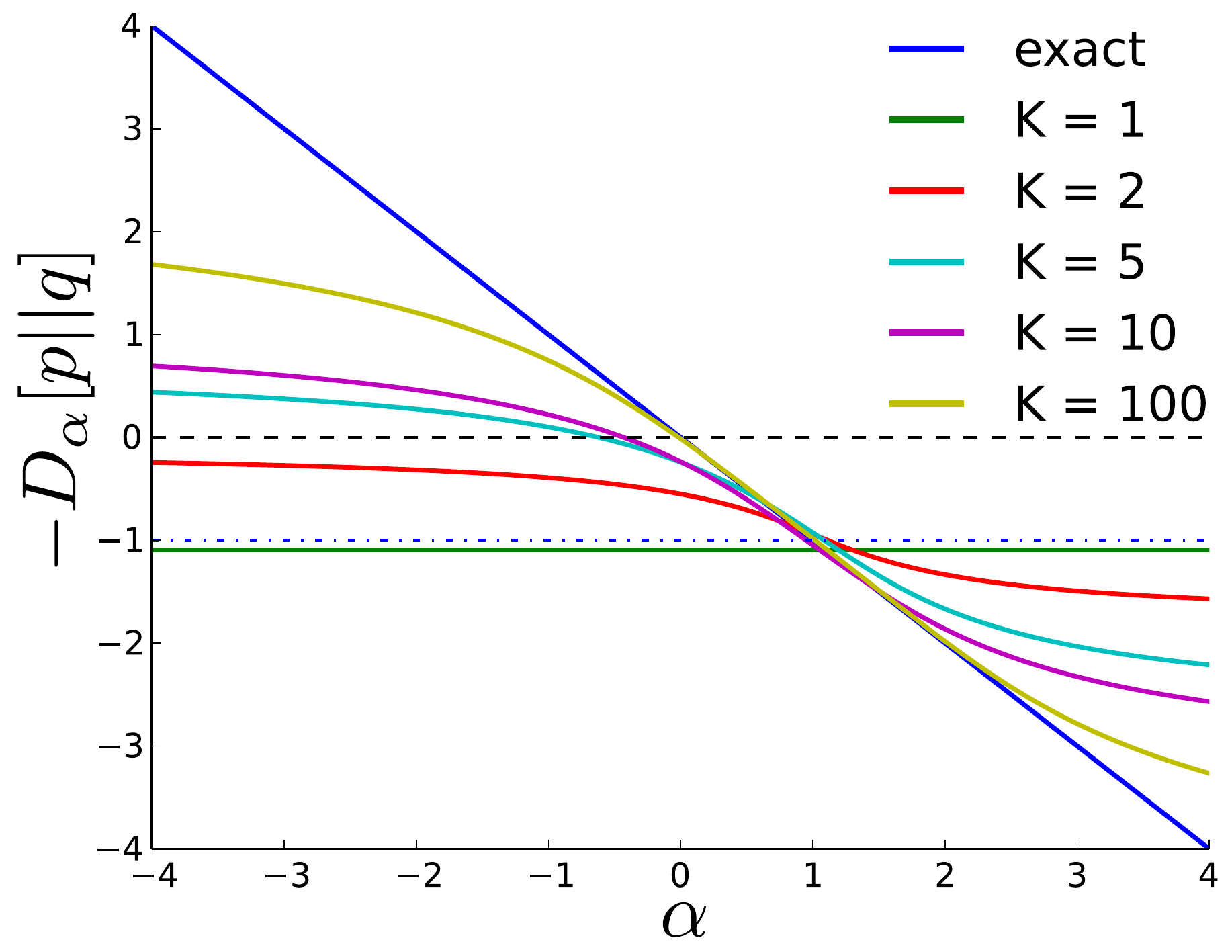}
 \label{fig:divergence_simulation}}
 \vspace{-0.1in}
 \caption{(a) An illustration for the bounding properties of MC approximations to the VR bounds. (b) The bias of the MC approximation. Best viewed in colour and see the main text for details.}
\end{figure*}

The intuition behind the theorems is visualised in Figure \ref{fig:divergence_sampling}. By definition, the exact VR bound is a lower-bound or upper-bound of $\log p(\bm{x})$ when $\alpha > 0$ or $\alpha < 0$, respectively. However the MC approximation $\mathbb{E}[\hat{\mathcal{L}}_{\alpha, K}]$ biases the estimate towards $\mathcal{L}_{\text{VI}}$, where the approximation quality can be improved using more samples. Thus for finite samples and under mild conditions, negative alpha values can potentially be used to improve the accuracy of the approximation, at the cost of losing the upper-bound guarantee.
Figure \ref{fig:divergence_simulation} shows an empirical evaluation by computing the exact and the MC approximation of the R{\'e}nyi divergences. In this example $p$, $q$ are 2-D Gaussian distributions with $\bm{\mu}_p = [0, 0]$, $\bm{\mu}_q = [1, 1]$ and $\bm{\Sigma}_p = \bm{\Sigma}_q = \bm{I}$. The sampling procedure is repeated 200 times to estimate the expectation. Clearly for $K = 1$ it is equivalent to an unbiased estimate of the KL-divergence for all $\alpha$ (even though now the estimation is biased for $\mathrm{D}_{\alpha}$). For $K > 1$ and $\alpha < 1$, the MC method under-estimates the VR bound, and the bias decreases with increasing $K$. For $\alpha > 1$ the inequality is reversed also as predicted.

\subsection{Unified implementation with the reparameterization trick}

Readers may have noticed that $\mathcal{L}_{\text{VI}}$ has a different form compared to $\mathcal{L}_{\alpha}$ with $\alpha \neq 1$. In this section we show how to unify the implementation for all finite $\alpha$ settings using the \emph{reparameterization trick} \cite{salimans:reparam, kingma:vae} as an example. This trick assumes the existence of the mapping $\bm{\theta} = g_{\bm{\phi}}(\bm{\epsilon})$, where the distribution of the noise term $\bm{\epsilon}$ satisfies $q(\bm{\theta}) d\bm{\theta} = p(\bm{\epsilon}) d\bm{\epsilon}$. Then the expectation of a function $F(\bm{\theta})$ over distribution $q(\bm{\theta})$ can be computed as
$\mathbb{E}_{q(\bm{\theta})} [F(\bm{\theta})] = \mathbb{E}_{p(\bm{\epsilon})} [F(g_{\bm{\phi}}(\bm{\epsilon}))].$
One prevalent example is the Gaussian reparameterization: $\bm{\theta} \sim \mathcal{N}(\bm{\mu}, \Sigma) \Rightarrow \bm{\theta} = \bm{\mu} + \Sigma^{ \frac{1}{2} } \bm{\epsilon}, \bm{\epsilon} \sim \mathcal{N}(\bm{0}, I)$. 
Now we apply the reparameterization trick to the VR bound
\begin{equation}
\mathcal{L}_{\alpha}(q_{\bm{\phi}}; \bm{x}) = \frac{1}{1 - \alpha} \log \mathbb{E}_{\bm{\epsilon}} \left[\left( \frac{p(g_{\bm{\phi}}(\bm{\epsilon}) , \bm{x})}{q(g_{\bm{\phi}}(\bm{\epsilon}))} \right)^{1 - \alpha} \right].
\end{equation}
Then the gradient of the VR bound w.r.t.~$\bm{\phi}$ is (similar for $\bm{\varphi}$, see appendix for derivation)
\begin{equation}
\nabla_{\bm{\phi}} \mathcal{L}_{\alpha}(q_{\bm{\phi}}; \bm{x}) 
= \mathbb{E}_{\bm{\epsilon}} \left[ w_{\alpha}(\bm{\epsilon}; \bm{\phi}, \bm{x}) \nabla_{\bm{\phi}} \log \frac{p(g_{\bm{\phi}}(\bm{\epsilon}), \bm{x})}{q(g_{\bm{\phi}}(\bm{\epsilon}))} \right],
\label{eq:gradient_reparam}
\end{equation}
where $w_{\alpha}(\bm{\epsilon}; \bm{\phi}, \bm{x}) = \left( \frac{p(g_{\bm{\phi}}(\bm{\epsilon}), \bm{x})}{q(g_{\bm{\phi}}(\bm{\epsilon}))} \right)^{1 - \alpha} \bigg/ \mathbb{E}_{\bm{\epsilon}} \left[ \left( \frac{p(g_{\bm{\phi}}(\bm{\epsilon}), \bm{x})}{q(g_{\bm{\phi}}(\bm{\epsilon}))} \right)^{1 - \alpha} \right]$ denotes the normalised importance weight. 
One can show that this recovers the the stochastic gradients of $\mathcal{L}_{\text{VI}}$ by setting $\alpha = 1$ in (\ref{eq:gradient_reparam}) since now $w_{1}(\bm{\epsilon}; \bm{\phi}, \bm{x}) = 1$, which means the resulting algorithm unifies the computation for all finite $\alpha$ settings. For MC approximations, we use $K$ samples to approximately compute the weight $\hat{w}_{\alpha, k}(\bm{\epsilon}_k; \bm{\phi}, \bm{x}) \propto \left( \frac{p(g_{\bm{\phi}}(\bm{\epsilon}_k), \bm{x})}{q(g_{\bm{\phi}}(\bm{\epsilon}_k))} \right)^{1 - \alpha}$, $k = 1, ..., K$, and the stochastic gradient becomes
\begin{equation}
\nabla_{\bm{\phi}} \hat{\mathcal{L}}_{\alpha, K}(q_{\bm{\phi}}; \bm{x}) 
= \sum_{k=1}^K \left[ \hat{w}_{\alpha, k}(\bm{\epsilon}_k; \bm{\phi}, \bm{x}) \nabla_{\bm{\phi}} \log \frac{p(g_{\bm{\phi}}(\bm{\epsilon}_k), \bm{x})}{q(g_{\bm{\phi}}(\bm{\epsilon}_k))} \right].
\end{equation}
When $\alpha = 1$, $\hat{w}_{1, k}(\bm{\epsilon}_k; \bm{\phi}, \bm{x}) = 1/K$, and it recovers the stochastic gradient VI method \cite{kingma:vae}.

\begin{figure}[!t]
\begin{minipage}[b]{0.5\linewidth}
\centering
\begin{algorithm}[H] 
\caption{One gradient step for VR-$\alpha$/VR-max with single backward pass. Here $\hat{w}(\bm{\epsilon}_k; \bm{x})$ short-hands $\hat{w}_{0, k}(\bm{\epsilon}_k; \bm{\phi}, \bm{x})$ in the main text.} \small
\label{alg:vr_max} 
\begin{algorithmic}[1] 
	\STATE given the current datapoint $\bm{x}$, sample \\$\bm{\epsilon}_1, ..., \bm{\epsilon}_K \sim p(\bm{\epsilon})$
	\STATE for $k = 1, ..., K$, compute the unnormalised weight
	$$\log \hat{w}(\bm{\epsilon}_k; \bm{x}) = \log p(g_{\bm{\phi}}(\bm{\epsilon}_k), \bm{x}) - \log q(g_{\bm{\phi}}(\bm{\epsilon}_k)|\bm{x})$$
	\STATE choose the sample $\bm{\epsilon}_{j}$ to back-propagate: \\
	if $|\alpha| < \infty$: $j \sim p_k$ where $p_k \propto \hat{w}(\bm{\epsilon}_k; \bm{x})^{1 - \alpha}$ \\
	if $\alpha = -\infty$: $j = \argmax_{k} \log \hat{w}(\bm{\epsilon}_k; \bm{x})$ 
	\STATE return the gradients $\nabla_{\bm{\phi}} \log \hat{w}(\bm{\epsilon}_{j}; \bm{x})$
\end{algorithmic}
\end{algorithm}
\end{minipage}
\hfill
\begin{minipage}[b]{0.4\linewidth}
\centering
 \includegraphics[width=0.9\linewidth]{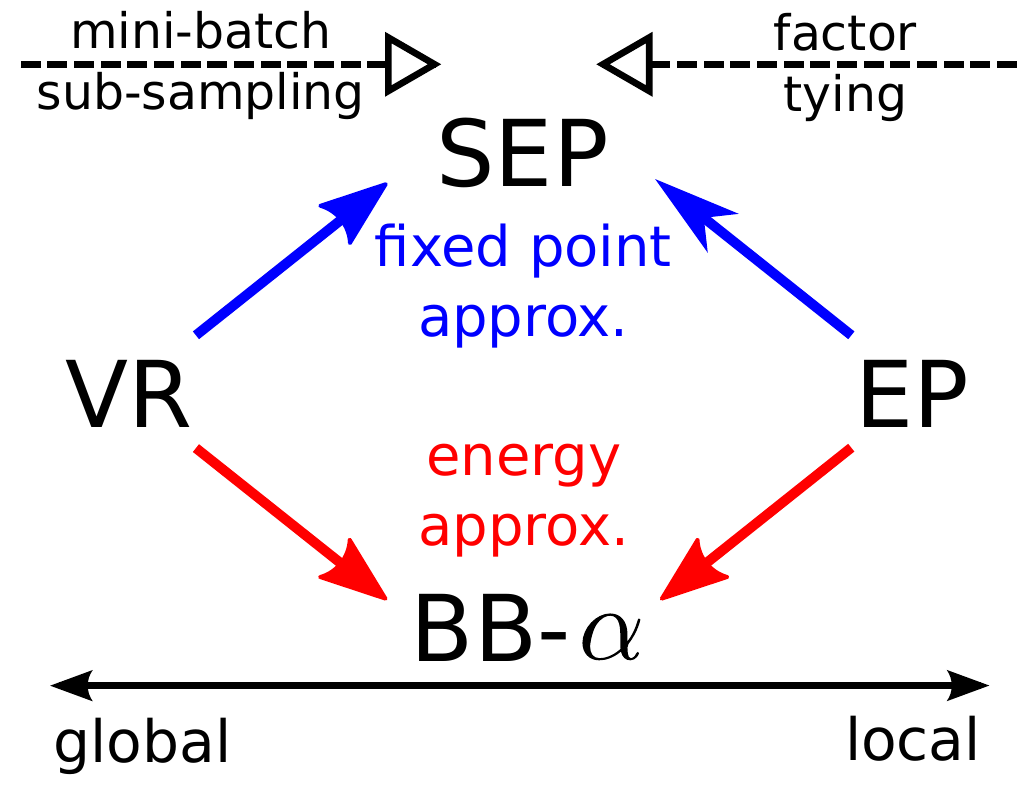} 
 \vspace{-0.1in}
 \captionof{figure}{Connecting local and global divergence minimisation.}
 \label{fig:vr_ep_relationship}
\end{minipage}
%
\end{figure}

To speed-up learning \cite{burda:iwae} suggested back-propagating only one sample $\bm{\epsilon}_j$ with $j \sim p_j = \hat{w}_{\alpha, j}$, which can be easily extended to our framework. Importantly, the use of different $\alpha < 1$ indicates the degree of emphasis placed upon locations where the approximation $q$ under-estimates $p$, and in the extreme case $\alpha \rightarrow -\infty$, the algorithm chooses the sample that has the \emph{maximum} unnormalised importance weight. 
We name this approach \emph{VR-max} and summarise it and the general case in Algorithm \ref{alg:vr_max}. 
Note that VR-max (and VR-$\alpha$ with $\alpha < 0$ and MC approximations) does \emph{not} minimise $\mathrm{D}_{1-\alpha}[p||q]$. It is true that $\mathcal{L}_{\alpha} \geq \log p(\bm{x})$ for negative $\alpha$ values. However Corollary \ref{thm:alpha_k_existence} suggests that the tightest MC approximation for given $K$ has non-positive $\alpha_K$ value, or might not even exist. Furthermore $\alpha_K$ becomes more negative as the mismatch between $q$ and $p$ increases, e.g.~VAE uses a uni-modal $q$ distribution to approximate the typically multi-modal exact posterior.

\subsection{Stochastic approximation for large-scale learning}
\label{sec:large_scale_learning}
VR bounds can also be applied to full Bayesian inference with posterior approximation. However for large datasets full batch learning is very inefficient. Mini-batch training is non-trivial here since the VR bound cannot be represented by the expectation on a datapoint-wise loss, except when $\alpha = 1$. This section introduces two proposals for mini-batch training, and interestingly, this recovers two existing algorithms that were motivated from a different perspective.
In the following we define the ``average likelihood'' $\bar{f}_{\mathcal{D}}(\bm{\theta}) = [\prod_{n=1}^N p(\bm{x}_n|\bm{\theta})]^{\frac{1}{N}}$. Hence the joint distribution can be rewritten as $p(\bm{\theta}, \mathcal{D}) = p_0(\bm{\theta}) \bar{f}_{\mathcal{D}}(\bm{\theta})^N$. Also for a mini-batch of $M$ datapoints $\mathcal{S} = \{\bm{x}_{n_1}, ..., \bm{x}_{n_M} \} \sim \mathcal{D}$, we define the ``subset average likelihood'' $\bar{f}_{\mathcal{S}}(\bm{\theta}) = [\prod_{m=1}^M p(\bm{x}_{n_m}|\bm{\theta})]^{\frac{1}{M}}$.

The first proposal considers \emph{fixed point approximations} with mini-batch sub-sampling. It first derives the fixed point conditions for the variational parameters (e.g.~the natural parameters of $q$) using the exact VR bound (\ref{eq:alpha_vi}), then design an iterative algorithm using those fixed point equations, but with $\bar{f}_{\mathcal{D}}(\bm{\theta})$ replaced by $\bar{f}_{\mathcal{S}}(\bm{\theta})$.
The second proposal also applies this subset average likelihood approximation idea, but directly to the VR bound (\ref{eq:alpha_vi}) (so this approach is named \emph{energy approximation}):
\begin{equation}
\label{eq:alpha_vi_approx}
\tilde{\mathcal{L}}_{\alpha}(q; \mathcal{S}) 
	= \frac{1}{1 - \alpha} \log \mathbb{E}_{q} \left[ \left( \frac{p_0(\bm{\theta}) \bar{f}_{\mathcal{S}}(\bm{\theta})^N} {q(\bm{\theta})} \right)^{1 - \alpha} \right].
\end{equation}
In the appendix we demonstrate with detailed derivations that fixed point approximation returns Stochastic EP (SEP) \cite{li:sep}, and black box alpha (BB-$\alpha$) \cite{hernandez-lobato:bb-alpha} corresponds to energy approximation. Both algorithms were originally proposed to approximate (power) EP \cite{minka:ep, minka:powerep}, which usually minimises $\alpha$-divergences \emph{locally}, and considers $M=1$, $\alpha \in [1 - 1/N, 1)$ and exponential family distributions. These approximations were done by factor tying, which significantly reduces the memory overhead of full EP and makes both SEP and BB-$\alpha$ scalable to large datasets just as SVI. The new derivation derivation provides a theoretical justification from energy perspective, and also sheds lights on the connections between \emph{local} and \emph{global} divergence minimisations as depicted in Figure \ref{fig:vr_ep_relationship}. Note that all these methods recover SVI when $\alpha \rightarrow 1$, in which global and local divergence minimisation are equivalent. Also these results suggest that recent attempts of distributed posterior approximation (by carving up the dataset into pieces with $M > 1$ \cite{gelman:dep, xu:sms}) can be extended to both SEP and BB-$\alpha$.

Monte Carlo methods can also be applied to both proposals. For SEP the moment computation can be approximated with MCMC \cite{gelman:dep, xu:sms}. For BB-$\alpha$ one can show in the same way as to prove Theorem \ref{thm:sampling_bound} that simple MC approximation in expectation lower-bounds the BB-$\alpha$ energy when $\alpha \leq 1$. In general it is also an open question how to choose $\alpha$ for given the mini-batch size $M$ and the number of samples $K$, but there is evidence that intermediate $\alpha$ values can be superior \citep{bui:dgp, depeweg:bnn_rl}.

\section{Experiments}
We evaluate the VR bound methods on Bayesian neural networks and variational auto-encoders. All the experiments used the ADAM optimizer \cite{kingma:adam}, and the detailed experimental set-up (batch size, learning rate, etc.) can be found in the appendix. The implementation of all the experiments in Python is released at \url{https://github.com/YingzhenLi/VRbound}.

\subsection{Bayesian neural network}

\begin{figure}[t]
 \centering
 \includegraphics[width=1.0\linewidth]{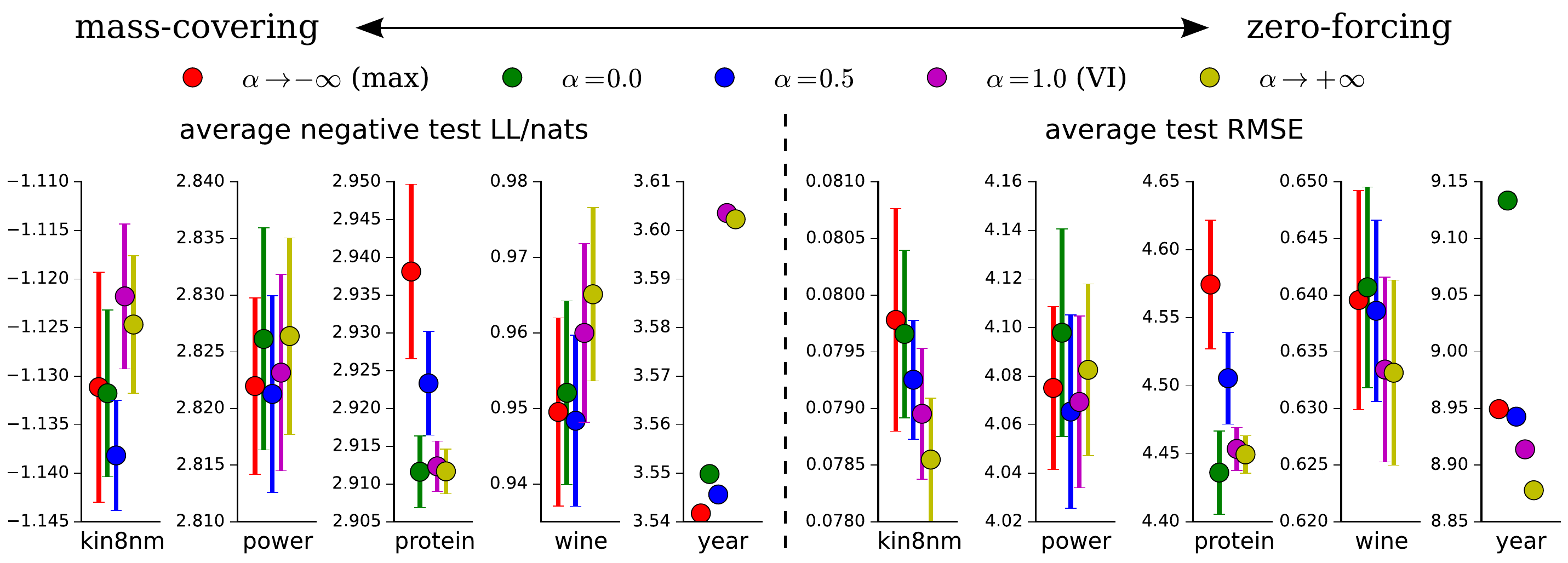} 
 \vspace{-0.15in}
 \caption{Test LL and RMSE results for Bayesian neural network regression. The lower the better.}
 \label{fig:bnn_results}
\end{figure}

The first experiment considers Bayesian neural network regression. The datasets are collected from the UCI dataset repository.\footnote{\url{http://archive.ics.uci.edu/ml/datasets.html}} The model is a single-layer neural network with 50 hidden units (ReLUs) for all datasets except Protein and Year (100 units). We use a Gaussian prior $\bm{\theta} \sim \mathcal{N}(\bm{\theta}; \bm{0}, \bm{I})$ for the network weights and Gaussian approximation to the true posterior $q(\bm{\theta}) = \mathcal{N}(\bm{\theta}; \bm{\mu}_q, diag(\bm{\sigma}_q))$. We follow the toy example in Section \ref{sec:vr_bound} to consider $\alpha \in \{-\infty, 0.0, 0.5, 1.0, +\infty \}$ in order to examine the effect of mass-covering/zero-forcing behaviour. Stochastic optimisation uses the energy approximation proposed in Section \ref{sec:large_scale_learning}. MC approximation is also deployed to compute the energy function, in which $K=100, 10$ is used for small and large datasets (Protein and Year), respectively.

We summarise the test negative log-likelihood (LL) and RMSE with standard error (across different random splits except for Year) for selected datasets in Figure \ref{fig:bnn_results}, where the full results are provided in the appendix. These results indicate that for posterior approximation problems, the optimal $\alpha$ may vary for different datasets. Also the MC approximation complicates the selection of $\alpha$ (see appendix). Future work should develop algorithms to automatically select the best $\alpha$ values, although a naive approach could use validation sets. We observed two major trends that zero-forcing/mode-seeking methods tend to focus on improving the predictive error, while mass-covering methods returns better calibrated uncertainty estimate and better test log-likelihood. In particular VI returns lower test log-likelihood for most of the datasets. Furthermore, $\alpha = 0.5$ produced overall good results for both test LL and RMSE, possibly because the skew symmetry is centred at $\alpha = 0.5$ and the corresponding divergence is the only symmetric distance measure in the family.

\subsection{Variational auto-encoder}

The second experiments considers variational auto-encoders for unsupervised learning. We mainly compare three approaches: VAE ($\alpha = 1.0$), IWAE ($\alpha = 0$), and VR-max ($\alpha = -\infty$), which are implemented upon the publicly available code.\footnote{\url{https://github.com/yburda/iwae}}
Four datasets are considered: Frey Face (with 10-fold cross validation), Caltech 101 Silhouettes, MNIST and OMNIGLOT. The VAE model has $L=1, 2$ stochastic layers with deterministic layers stacked between, and the network architecture is detailed in the appendix. 
We reproduce the IWAE experiments to obtain a fair comparison, since the results in the original publication \cite{burda:iwae} mismatches those evaluated on the publicly available code. 

\begin{figure}[t]
\begin{minipage}{0.6\linewidth}

\centering
\captionof{table}{ Average Test log-likelihood. Results for VAE on MNIST and OMNIGLOT are collected from \cite{burda:iwae}.}
\label{tab:vae_ll}
\begin{tabular}{lccccc}
\toprule
\bf{Dataset}&$L$&$K$&\bf{ VAE }&\bf{ IWAE }&\bf{ VR-max }\\
\hline
Frey Face&1&5&1322.96&\bf{1380.30}&1377.40\\
($\pm$ std. err.)& & &$\pm$10.03&\bf{$\pm$4.60}&$\pm$4.59 \\
\hline
Caltech 101& 1& 5&-119.69&\bf{-117.89}&-118.01\\
Silhouettes&  &50&-119.61&-117.21&\bf{-117.10}\\
\hline
MNIST& 1&5 &-86.47&\bf{-85.41}&-85.42\\
     &  &50&-86.35&\bf{-84.80}&-84.81\\
     & 2&5 &-85.01&\bf{-83.92}&-84.04\\
     &  &50&-84.78&\bf{-83.05}&-83.44\\
\hline
OMNIGLOT & 1&5 &-107.62&\bf{-106.30}&-106.33\\
		 & 1&50&-107.80&\bf{-104.68}&-105.05\\
		 & 2&5 &-106.31&\bf{-104.64}&-104.71\\
		 & 2&50&-106.30&\bf{-103.25}&-103.72\\
\bottomrule
\end{tabular}

\end{minipage}
\hfill
\begin{minipage}{0.35\linewidth}
\centering
 \includegraphics[width=1.00\linewidth]{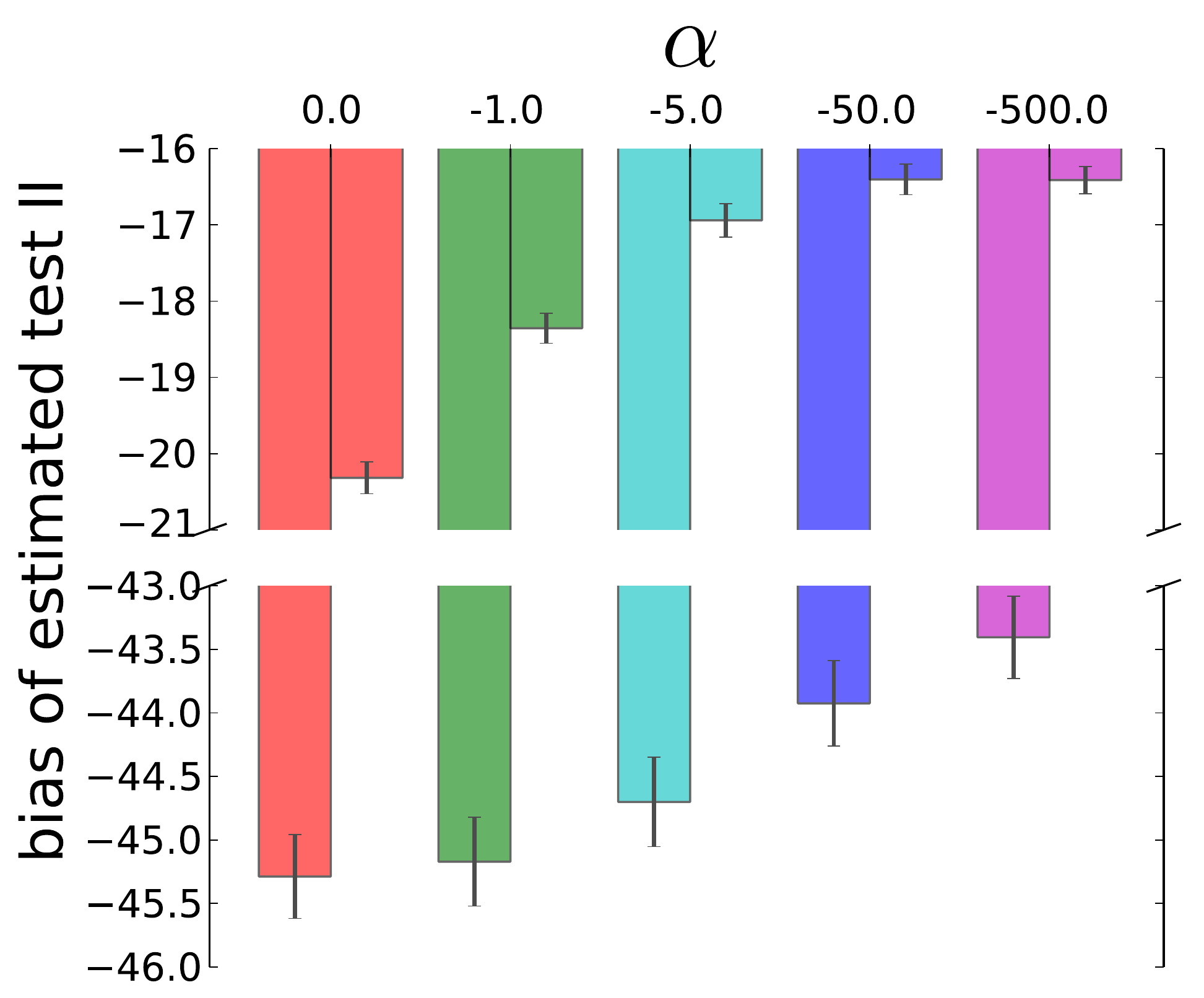} 
 \vspace{-0.1in}
 \captionof{figure}{Bias of sampling approximation to. Results for $K=5, 50$ samples are shown on the left and right, respectively.}
 \label{fig:vae_bias}
\end{minipage}
\vspace{-0.1in}
\end{figure}
\begin{figure*}[!ht]
 \vspace{-0.1in}
 \centering
 \subfigure[Log of ratio $R = w_{max} / (1 - w_{max})$]{
 \includegraphics[width=0.37\linewidth]{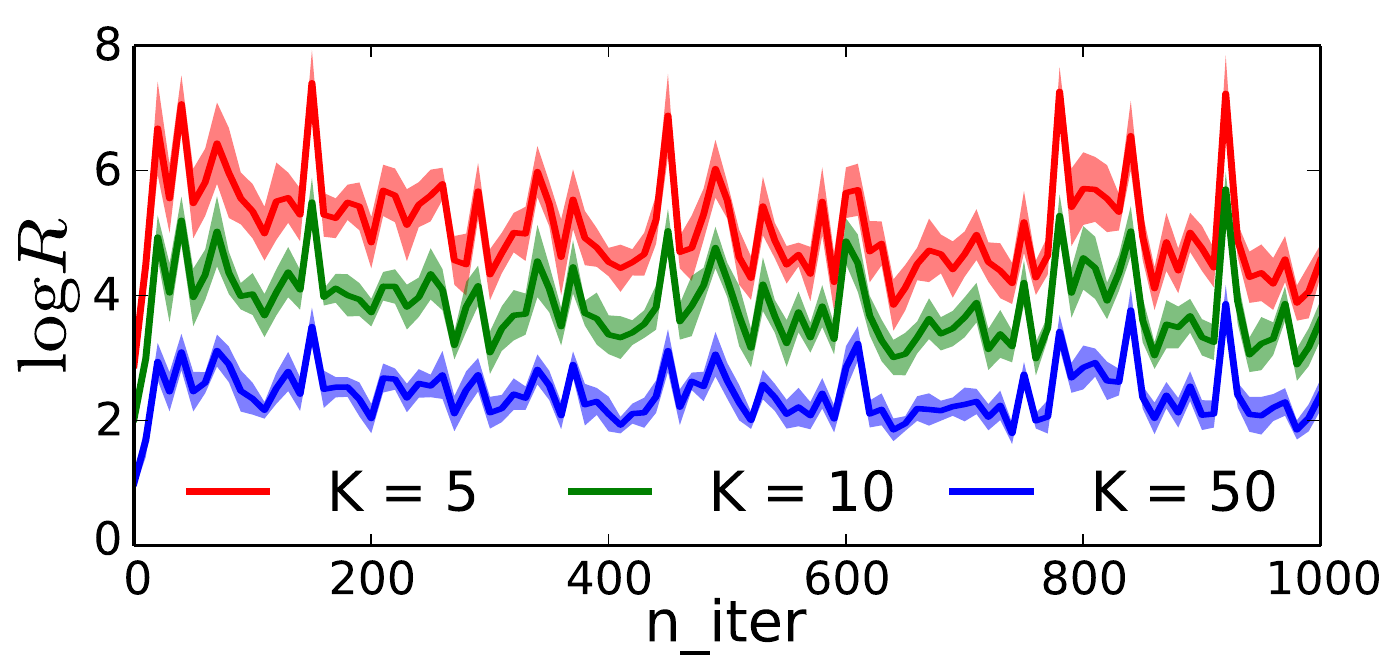}
 \label{fig:weight_ratio}}
 \hspace{0.4in}
 \subfigure[Weights of samples.]{
 \includegraphics[width=0.37\linewidth]{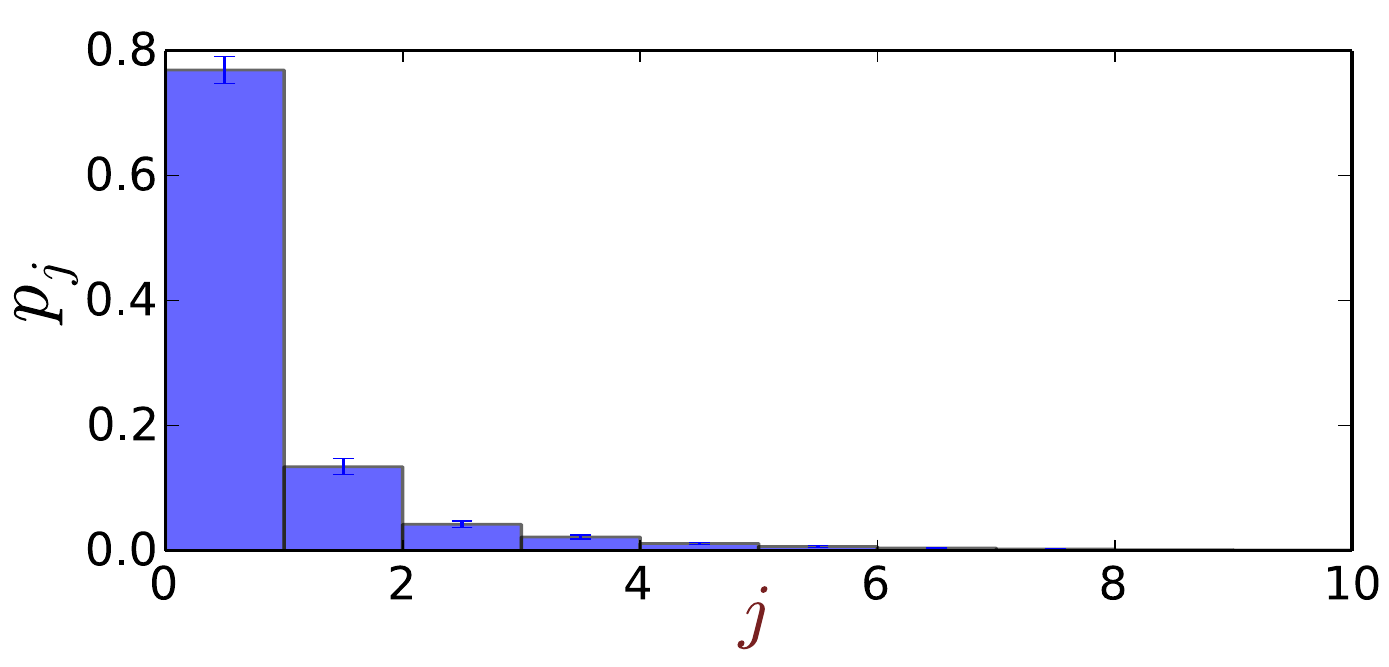}
 \label{fig:weight_distribution}}
\vspace{-0.1in}
 \caption{Importance weights during training, see main text for details. Best viewed in colour.}
 \vspace{-0.1in}
\end{figure*}

We report test log-likelihood results in Table \ref{tab:vae_ll} by computing $\log p(\bm{x}) \approx \hat{\mathcal{L}}_{0, 5000}(q; \bm{x})$ following \cite{burda:iwae}. We also present some samples from the trained models in the appendix. 
Overall VR-max is almost indistinguishable from IWAE. Other positive alpha settings (e.g.~$\alpha = 0.5$) return worse results, e.g. $1374.64\pm5.62$ for Frey Face and $-85.50$ for MNIST with $\alpha = 0.5$, $L=1$ and $K=5$. These worse results for $\alpha > 0$ indicate the preference of getting tighter approximations to the likelihood function for MLE problems. Small negative $\alpha$ values (e.g.~$\alpha = -1.0, -2.0$) returns better results on different splits of the Frey Face data, and overall the best $\alpha$ value is dataset-specific. 

VR-max's success might be explained by the tightness of the bound. To evaluate this, we compute the VR bounds on $100$ test datapoints using the 1-layer VAE trained on Frey Face, with $K=\{5, 50\}$ and $\alpha \in \{0, -1, -5, -50, -500 \}$. Figure \ref{fig:vae_bias} presents the estimated gap $\hat{\mathcal{L}}_{\alpha, K} - \hat{\mathcal{L}}_{0, 5000}$. The results indicates that $\hat{\mathcal{L}}_{\alpha, K}$ provides a lower-bound, and that gap is narrowed as $\alpha \rightarrow -\infty$. Also increasing $K$ provides improvements. The standard error of estimation is almost constant for different $\alpha$ (with $K$ fixed), and is negligible when compared to the MC approximation bias.

Another explanation for VR-max's success is that, the sample with the largest normalised importance weight $w_{max}$ dominates the contributions of all the gradients.
This is confirmed by tracking $R = \frac{w_{max}}{1 - w_{max}}$ during training on Frey Face (Figure \ref{fig:weight_ratio}). Also Figure \ref{fig:weight_distribution} shows the $10$ largest importance weights from $K=50$ samples in descending order, which exhibit an exponential decay behaviour, with the largest weight occupying more than $75\%$ of the probability mass. Hence VR-max provides a fast approximation to IWAE when tested on CPUs or multiple GPUs with high communication costs. Indeed our numpy implementation of VR-max achieves up to 3 times speed-up compared to IWAE (9.7s vs.~29.0s per epoch, tested on Frey Face data with $K = 50$ and batch size $M = 100$, CPU info: Intel Core i7-4930K CPU @ 3.40GHz). However this speed advantage is less significant when the gradients can be computed very efficiently on a single GPU.

\section{Conclusion}
We have introduced the variational R{\'e}nyi bound and an associated optimisation framework. We have shown the richness of the new family, not only by connecting to existing approaches including VI/VB, SEP, BB-$\alpha$, VAE and IWAE, but also by proposing the VR-max algorithm as a new special case. Empirical results on Bayesian neural networks and variational auto-encoders indicate that VR bound methods are widely applicable and can obtain state-of-the-art results.
Future work will focus on both experimental and theoretical sides. Theoretical work will study the interaction of the biases introduced by MC approximation and datapoint sub-sampling. A guide on choosing optimal $\alpha$ values are needed for practitioners when applying the framework to their applications.

\subsubsection*{Acknowledgements}
We thank the Cambridge MLG members and the reviewers for comments. YL thanks the Schlumberger Foundation FFTF fellowship. RET thanks EPSRC grants \# EP/M026957/1 and EP/L000776/1.

\subsubsection*{References}
\renewcommand{\section}[2]{}
\setlength{\bibsep}{0pt plus 0.3ex}
\bibliographystyle{ieeetr}
\small
\bibliography{alpha_vi}



\section*{Supplementary material}

# Rényi Divergence Variational Inference: Appendix

**Yingzhen Li**
University of Cambridge
Cambridge, CB2 1PZ, UK
yl494@cam.ac.uk

**Richard E. Turner**
Universityof Cambridge
Cambridge, CB2 1PZ, UK
ret26@cam.ac.uk

The appendix is organised as follows. Section A presents other existing definitions of $\alpha$-divergences. Section B provides the mathematical details for the Bayesian linear regression example. Section C provides the proofs for the main theoretical results. Section D briefly discusses the optimisation issues brought from the selection of $\alpha$ values and the number of MC samples $K$. Section E applies the reparametrization trick to the MC approximated bound, which leads to a unified implementation. Section F demonstrates the connections between the proposed sub-sampling approximation and existing algorithms (SEP [1] and BB-$\alpha$ [2]). Section G provides detailed experimental set-up and further results for the tests considered in the main text.

## A  Other $\alpha$-divergence definitions

Here we include some existing $\alpha$-divergence definitions other than Rényi's.

- Amari's $\alpha$-divergence [3]

$$\mathrm{D}_\alpha[p||q] = \frac{4}{1-\alpha^2}\left(1 - \int p(\boldsymbol{\theta})^{\frac{1+\alpha}{2}} q(\boldsymbol{\theta})^{\frac{1-\alpha}{2}} d\boldsymbol{\theta}\right).$$

- Tsallis's $\alpha$-divergence [4]

$$\mathrm{D}_\alpha[p||q] = \frac{1}{\alpha-1}\left(\int p(\boldsymbol{\theta})^\alpha q(\boldsymbol{\theta})^{1-\alpha} d\boldsymbol{\theta} - 1\right).$$

Consider the problem of posterior approximation by minimising an $\alpha$-divergence. When the approximate posterior $q$ has an exponential family form, minimising $\mathrm{D}_\alpha[p||q]$, no matter which definition above is used (although may use different alpha), requires moment matching to the tilted distribution $\tilde{p}_\alpha(\boldsymbol{\theta}) \propto p(\boldsymbol{\theta})^\alpha q(\boldsymbol{\theta})^{1-\alpha}$. In the EP literature Amari's definition is often discussed. We focus on Rényi's definition in the main text simply because $\mathrm{D}_\alpha[q(\boldsymbol{\theta})||p(\boldsymbol{\theta}|\mathcal{D})]$ using Rényi's definition contains $\log p(\mathcal{D})$ that can be cancelled in the same way as VI is derived.

## B  A mean-field approximation example

We present the mean-field approximation method for the VR bound family, with Bayesian linear regression as an illustrating example. Recall the VR bound for $\alpha \neq 1$:

$$\mathcal{L}_\alpha(q; \mathcal{D}) := \frac{1}{1-\alpha} \log \mathbb{E}_q\left[\left(\frac{p(\boldsymbol{\theta}, \mathcal{D})}{q(\boldsymbol{\theta})}\right)^{1-\alpha}\right], \tag{1}$$

where the $q$ distribution is factorised over the components of $\boldsymbol{\theta} = (\theta_1, ..., \theta_d)$: $q(\boldsymbol{\theta}) = \prod_i q(\theta_i)$. In the following we denote $q_j = q(\theta_j)$ to reduce notational clutter, and re-write the VR bound as

$$\mathcal{L}_\alpha(q; \mathcal{D}) = \frac{1}{1-\alpha} \log \int \prod_i q_i \left(\frac{p(\boldsymbol{\theta}, \mathcal{D})}{\prod_i q_i}\right)^{1-\alpha} d\boldsymbol{\theta}$$

$$= \frac{1}{1-\alpha} \log \int q_j^\alpha \left(\int \prod_{i \neq j} q_i \left(\frac{p(\boldsymbol{\theta}, \mathcal{D})}{\prod_{i \neq j} q_i}\right)^{1-\alpha} d\boldsymbol{\theta}_{i \neq j}\right) d\theta_j$$

$$:= \frac{1}{1-\alpha} \log \int q_j^\alpha \tilde{p}_j^{1-\alpha} d\theta_j + \text{const},$$

where $\tilde{p}_j$ denote the "marginal" distribution satisfying

$$\log \tilde{p}_j = \frac{1}{1-\alpha} \log \int \prod_{i \neq j} q_i \left(\frac{p(\boldsymbol{\theta}, \mathcal{D})}{\prod_{i \neq j} q_i}\right)^{1-\alpha} d\boldsymbol{\theta}_{i \neq j} + \text{const}.$$

Now maximising the VR bound (when $\alpha > 0$, and for $\alpha < 0$ we minimise the bound) is equivalent to minimising $\text{D}_\alpha[q_j||\tilde{p}_j]$ (for $\alpha > 0$, and when $\alpha < 0$ we minimise $\text{D}_{1-\alpha}[\tilde{p}_j||q_j]$), which means $\log q_j = \log \tilde{p}_j + \text{const}$. One can verify that when $\alpha \to 1$ it recovers the traditional variational mean-field approximation

$$\lim_{\alpha \to 1} q_j = \int \prod_{i \neq j} q_i \log p(\boldsymbol{\theta}, \mathcal{D}) d\boldsymbol{\theta}_{i \neq j} + \text{const},$$

and when $\alpha \to 0$ it returns the exact marginal of the posterior distribution $\lim_{\alpha \to 0} q_j = p(\theta_j|\mathcal{D})$.

Now consider Bayesian linear regression with 2-D input $\boldsymbol{x}$ and 1-D output $y$, as an example:

$$\boldsymbol{\theta} \sim \mathcal{N}(\boldsymbol{\theta}; \boldsymbol{\mu}_0, \boldsymbol{\Lambda}_0^{-1}), \quad y|\boldsymbol{x} \sim \mathcal{N}(y; \boldsymbol{\theta}^T \boldsymbol{x}, \sigma^2).$$

Given the observations $\mathcal{D} = \{\boldsymbol{x}_n, y_n\}$, the posterior distribution of $\boldsymbol{\theta}$ can be computed analytically as $p(\boldsymbol{\theta}|\mathcal{D}) = \mathcal{N}(\boldsymbol{\theta}; \boldsymbol{\mu}, \boldsymbol{\Lambda}^{-1})$ with $\boldsymbol{\Lambda} = \boldsymbol{\Lambda}_0 + \frac{1}{\sigma^2} \sum_n \boldsymbol{x}_n \boldsymbol{x}_n^T$ and $\boldsymbol{\Lambda}\boldsymbol{\mu} = \boldsymbol{\Lambda}_0 \boldsymbol{\mu}_0 + \frac{1}{\sigma^2} \sum_n y_n \boldsymbol{x}_n$. To see how the mean-field approach work we explicitly write down the elements of the posterior parameters

$$\boldsymbol{\mu} = \begin{pmatrix} \mu_1 \\ \mu_2 \end{pmatrix}, \quad \boldsymbol{\Lambda} = \begin{pmatrix} \Lambda_{11} & \Lambda_{12} \\ \Lambda_{21} & \Lambda_{22} \end{pmatrix}, \quad \Lambda_{12} = \Lambda_{21},$$

and define $q_i = \mathcal{N}(\theta_i; m_i, \lambda_i^{-1})$ as a univariate Gaussian distribution. Then

$$\log q_1 = \frac{1}{1-\alpha} \log \int q_2(\theta_2) \left(\frac{p(\boldsymbol{\theta}, \mathcal{D})}{q_2(\theta_2)}\right)^{1-\alpha} d\theta_2 + \text{const}$$

$$= \frac{1}{1-\alpha} \log \int \exp\left[-\frac{1-\alpha}{2}(\boldsymbol{\theta} - \boldsymbol{\mu})^T \boldsymbol{\Lambda}(\boldsymbol{\theta} - \boldsymbol{\mu}) - \frac{\alpha}{2}\lambda_2(\theta_2 - m_2)^2\right] d\theta_2 + \text{const}$$

$$= \frac{1}{1-\alpha} \log \int \mathcal{N}(\boldsymbol{\theta}; \boldsymbol{\mu}, \tilde{\boldsymbol{\Sigma}}) d\theta_2 + \text{const}$$

$$= \log \mathcal{N}(\theta_1; m_1, \lambda^{-1}) + \text{const}$$

where the new mean $m_1$ and the precision $\lambda_1$ satisfies

$$m_1 = \mu_1 + C_1(\mu_2 - m_2), \quad C_1 = \frac{\alpha \lambda_2 \Lambda_{12}}{(1-\alpha)|\boldsymbol{\Lambda}| + \alpha \lambda_2 \Lambda_{11}},$$

$$\lambda_1 = \Lambda_{11} - (1-\alpha)\Lambda_{12}((1-\alpha)\Lambda_{22} + \alpha\lambda_2)^{-1}\Lambda_{21}.$$

One can derive the terms $m_2$ and $C_2$ for $q_2$ in the same way, and show that $\boldsymbol{m} = \boldsymbol{\mu}$ is the only stable fixed point of this iterative update. So we have $q_1 = \mathcal{N}(\theta_1; \mu_1, \lambda_1^{-1})$, and similarly $q_2 = \mathcal{N}(\theta_1; \mu_2, \lambda_2^{-1})$ with $\lambda_2 = \Lambda_{22} - (1-\alpha)\Lambda_{21}((1-\alpha)\Lambda_{11} + \alpha\lambda_1)^{-1}\Lambda_{12}$. In this example $\lambda_1, \lambda_2$ are feasible for all $\alpha$, and solving the fixed point equations, finally we have the stable fixed point as

$$\lambda_1 = \rho_\alpha \Lambda_{11}, \quad \lambda_2 = \rho_\alpha \Lambda_{22}, \quad \rho_\alpha = \frac{1}{2\alpha}\left[(2\alpha - 1) + \sqrt{1 - \frac{4\alpha(1-\alpha)\Lambda_{12}^2}{\Lambda_{11}\Lambda_{22}}}\right].$$

The other solution for the quadratic formula is eliminated since it violates the assumptions that $\lambda_1 > 0$ (when $0 < \alpha < 1$) and $|\mathcal{L}_\alpha| < +\infty$ (when $\alpha < 0$ or $\alpha > 1$, since it requires $|\alpha \text{diag}(\boldsymbol{\lambda}) + (1-\alpha)\boldsymbol{\Lambda}| > 0$). Thus the stable fixed point in this case is unique.

One can show that $\lim_{\alpha \to 1} \lambda_1 = \Lambda_{11}$, $\lim_{\alpha \to 0} \lambda_1 = \Lambda_{11} - \Lambda_{12}\Lambda_{22}^{-1}\Lambda_{21}$ and $\lim_{\alpha \to \pm\infty} \lambda_1 = \Lambda_{11} \pm |\Lambda_{12}|\sqrt{\Lambda_{11}\Lambda_{22}^{-1}}$ (similar results for $\lambda_2$). Also $\rho_\alpha$ is continuous and non-decreasing in $\alpha$. This means one can interpolate between mass-covering and zero-forcing behaviour by increasing $\alpha$ values. Moreover, notice that the limiting case $\alpha \to +\infty$ still returns uncertain estimates, although it is even more over-confident than VI. This is different from maximum a posteriori (MAP) which captures the mode but only returns a point estimate.

## C  Proofs of the main results

We provide the proofs of the theorems presented in section 4 of the main text.

### C.1  Proof of Theorem 2

*Proof.* 1) First we prove for $\alpha \leq 1$, $\mathbb{E}_{\{\boldsymbol{h}_k\}}[\hat{\mathcal{L}}_{\alpha,K}]$ is non-decreasing in $K$. It is straight forward to show the results holds for $\alpha = 1$. We follow the proof in [5] for fixed $\alpha < 1$. Let $K > 1$ and the subset of indices $I = \{i_1, ..., i_{K'}\} \subset \{1, ..., K\}, K' < K$ randomly sampled from integers 1 to $K$. Then for any $\alpha < 1$:

$$\mathbb{E}_{\{\boldsymbol{h}_k\}_{k=1}^K}[\hat{\mathcal{L}}_{\alpha,K}] = \frac{1}{1-\alpha}\mathbb{E}_{\{\boldsymbol{h}_k\}}\left[\log \frac{1}{K}\sum_{k=1}^K \left(\frac{p(\boldsymbol{h}_k, \boldsymbol{x})}{q(\boldsymbol{h}_k|\boldsymbol{x})}\right)^{1-\alpha}\right]$$

$$= \frac{1}{1-\alpha}\mathbb{E}_{\{\boldsymbol{h}_k\}}\left[\log \mathbb{E}_{I \subset \{1,...,K\}}\left[\frac{1}{K'}\sum_{k=1}^{K'} \left(\frac{p(\boldsymbol{h}_{i_k}, \boldsymbol{x})}{q(\boldsymbol{h}_{i_k}|\boldsymbol{x})}\right)^{1-\alpha}\right]\right]$$

$$\geq \frac{1}{1-\alpha}\mathbb{E}_{\{\boldsymbol{h}_k\}}\left[\mathbb{E}_{I \subset \{1,...,K\}}\left[\log \frac{1}{K'}\sum_{k=1}^{K'} \left(\frac{p(\boldsymbol{h}_{i_k}, \boldsymbol{x})}{q(\boldsymbol{h}_{i_k}|\boldsymbol{x})}\right)^{1-\alpha}\right]\right] \quad (\log x \text{ is concave})$$

$$= \frac{1}{1-\alpha}\mathbb{E}_{\{\boldsymbol{h}_k\}}\left[\log \frac{1}{K'}\sum_{k=1}^{K'} \left(\frac{p(\boldsymbol{h}_k, \boldsymbol{x})}{q(\boldsymbol{h}_k|\boldsymbol{x})}\right)^{1-\alpha}\right] = \mathbb{E}_{\{\boldsymbol{h}_k\}_{k=1}^{K'}}[\hat{\mathcal{L}}_{\alpha,K'}]$$

We used Jensen's inequality of logarithm for the lower-bounding result here. When $\alpha > 1$ we can proof similar result but with inequality reversed, simply because now $1 - \alpha < 0$.

2) Next we prove that, when $K \to \infty$ and $|\mathcal{L}_\alpha| < +\infty$, we have $\mathbb{E}_{\{\boldsymbol{h}_k\}_{k=1}^K}[\hat{\mathcal{L}}_{\alpha,K}] \to \mathcal{L}_\alpha$ if $\hat{\mathcal{L}}_{\alpha,K}$ is absolutely integrable wrt. $qd\mu = dQ$ for all $K \geq 1$ (in other words $\mathbb{E}_{\{\boldsymbol{h}_k\}_{k=1}^K}[|\hat{\mathcal{L}}_{\alpha,K}|] < +\infty$). We only prove it for $\alpha \leq 1$, and for $\alpha > 1$ it can be proved in a similar way. First we use Jensen's inequality again for all finite $K$:

$$\mathbb{E}_{\{\boldsymbol{h}_k\}_{k=1}^K}[\hat{\mathcal{L}}_{\alpha,K}] = \frac{1}{1-\alpha}\mathbb{E}_{\{\boldsymbol{h}_k\}}\left[\log \frac{1}{K}\sum_{k=1}^K \left(\frac{p(\boldsymbol{h}_k, \boldsymbol{x})}{q(\boldsymbol{h}_k|\boldsymbol{x})}\right)^{1-\alpha}\right]$$

$$\leq \frac{1}{1-\alpha}\log \mathbb{E}_{\{\boldsymbol{h}_k\}}\left[\frac{1}{K}\sum_{k=1}^K \left(\frac{p(\boldsymbol{h}_k, \boldsymbol{x})}{q(\boldsymbol{h}_k|\boldsymbol{x})}\right)^{1-\alpha}\right] = \mathcal{L}_\alpha.$$

This implies $\limsup_{K \to +\infty} \mathbb{E}_{\{\boldsymbol{h}_k\}_{k=1}^K}[\hat{\mathcal{L}}_{\alpha,K}] \leq \mathcal{L}_\alpha$.

Then as an intermediate result we prove $\hat{\mathcal{L}}_{\alpha,K} \to \mathcal{L}_\alpha$ almost surely when $K \to \infty$. For $\alpha \neq 1$, since function log is continuous we again swap the limit and logarithm:

$$\lim_{K \to +\infty} \frac{1}{1-\alpha}\log \frac{1}{K}\sum_{k=1}^K \left(\frac{p(\boldsymbol{h}_k, \boldsymbol{x})}{q(\boldsymbol{h}_k|\boldsymbol{x})}\right)^{1-\alpha} = \frac{1}{1-\alpha}\log \lim_{K \to +\infty} \frac{1}{K}\sum_{k=1}^K \left(\frac{p(\boldsymbol{h}_k, \boldsymbol{x})}{q(\boldsymbol{h}_k|\boldsymbol{x})}\right)^{1-\alpha}.$$

3

Now since we assume $|\mathcal{L}_\alpha| < +\infty$, this implies $\mathbb{E}_q\left[\left(\frac{p(\boldsymbol{h},\boldsymbol{x})}{q(\boldsymbol{h}|\boldsymbol{x})}\right)^{1-\alpha}\right]$ is finite. Also notice for all $\alpha$ values the ratio $p/q$ is non-negative. Thus by the strong law of large numbers we have

$$\lim_{K\to+\infty} \frac{1}{K}\sum_{k=1}^{K}\left(\frac{p(\boldsymbol{h}_k,\boldsymbol{x})}{q(\boldsymbol{h}_k|\boldsymbol{x})}\right)^{1-\alpha} = \mathbb{E}_{q(\boldsymbol{h}|\boldsymbol{x})}\left[\left(\frac{p(\boldsymbol{h},\boldsymbol{x})}{q(\boldsymbol{h}|\boldsymbol{x})}\right)^{1-\alpha}\right] \quad \text{a. s.},$$

then $\hat{\mathcal{L}}_{\alpha,K} \to \mathcal{L}_\alpha$ almost surely as $K \to +\infty$. When $\alpha = 1$ we can use similar method to prove $\lim_{K\to+\infty} \hat{\mathcal{L}}_{1,K} = \mathcal{L}_{\text{VI}}$ almost surely.

Finally, using the non-increasing in $\alpha$ result we will prove later we have $\hat{\mathcal{L}}_{\alpha,K} \geq \hat{\mathcal{L}}_{1,K}$. Thus we can apply Fatou's Lemma and obtain the following almost surely (notice $\mathbb{E}[\hat{\mathcal{L}}_{1,K}] = \mathcal{L}_{\text{VI}}$ for all $K$):

$$\mathcal{L}_\alpha - \mathcal{L}_{\text{VI}} = \mathbb{E}_{\{\boldsymbol{h}_k\}_{k=1}^K}[\lim_{K\to+\infty} \hat{\mathcal{L}}_{\alpha,K} - \hat{\mathcal{L}}_{1,K}]$$
$$\leq \liminf_{K\to+\infty} \mathbb{E}_{\{\boldsymbol{h}_k\}_{k=1}^K}[\hat{\mathcal{L}}_{\alpha,K} - \hat{\mathcal{L}}_{1,K}]$$
$$= \liminf_{K\to+\infty} \mathbb{E}_{\{\boldsymbol{h}_k\}_{k=1}^K}[\hat{\mathcal{L}}_{\alpha,K}] - \mathcal{L}_{\text{VI}}.$$

Combining with the supremum bound, we have $\mathbb{E}_{\{\boldsymbol{h}_k\}_{k=1}^K}[\hat{\mathcal{L}}_{\alpha,K}] \to \mathcal{L}_\alpha$ when $K$ goes to infinity. For $\alpha > 1$ we use Jensen's inequality to bound the limit infimum and the non-increasing property in $\alpha$ to bound the limit supremum. Thus the convergence result holds for all $\alpha \in \{\alpha : |\mathcal{L}_\alpha| < +\infty\}$.

3) $\mathbb{E}[\hat{\mathcal{L}}_{\alpha,K}]$ is non-increasing in $\alpha$: since expectation preserves monotonicity, it is sufficient to prove the result for $\hat{\mathcal{L}}_{\alpha,K}$. This can be proved in similar way as Theorem 3 and 39 in [6], and we include the prove here for completeness. Notice that for $\alpha < \beta$ function $x^{\frac{1-\alpha}{1-\beta}}$ defined on $x > 0$ is convex when $\alpha < 1$ and concave when $\alpha > 1$. So applying Jensen's inequality:

$$\hat{\mathcal{L}}_{\alpha,K} = \frac{1}{1-\alpha}\log\frac{1}{K}\sum_{k=1}^{K}\left(\frac{p(\boldsymbol{h}_k,\boldsymbol{x})}{q(\boldsymbol{h}_k|\boldsymbol{x})}\right)^{1-\alpha} = \frac{1}{1-\alpha}\log\frac{1}{K}\sum_{k=1}^{K}\left(\left(\frac{p(\boldsymbol{h}_k,\boldsymbol{x})}{q(\boldsymbol{h}_k|\boldsymbol{x})}\right)^{1-\beta}\right)^{\frac{1-\alpha}{1-\beta}}$$
$$\geq \frac{1}{1-\alpha}\log\left(\frac{1}{K}\sum_{k=1}^{K}\left(\frac{p(\boldsymbol{h}_k,\boldsymbol{x})}{q(\boldsymbol{h}_k|\boldsymbol{x})}\right)^{1-\beta}\right)^{\frac{1-\alpha}{1-\beta}} = \hat{\mathcal{L}}_{\beta,K}.$$

Continuity in $\alpha$: First we show $\hat{\mathcal{L}}_{\alpha,K}$ is continuous in $\alpha$ when $p(\boldsymbol{h}_k,\boldsymbol{x}) \neq 0$ for $\boldsymbol{h}_k \sim q$. For $\alpha \neq 0, 1, \infty$ and for any sequence $\{\alpha_n\} \to \alpha$ it is sufficient to show that

$$\lim_{n\to\infty} \log\frac{1}{K}\sum_k q(\boldsymbol{h}_k|\boldsymbol{x})^{\alpha_n} p(\boldsymbol{h}_k,\boldsymbol{x})^{1-\alpha_n}$$
$$= \log\lim_{n\to\infty} \frac{1}{K}\sum_k q(\boldsymbol{h}_k|\boldsymbol{x})^{\alpha_n} p(\boldsymbol{h}_k,\boldsymbol{x})^{1-\alpha_n} \quad (\log x \text{ is a continuous function})$$
$$= \log\frac{1}{K}\sum_k \lim_{n\to\infty} q(\boldsymbol{h}_k|\boldsymbol{x})^{\alpha_n} p(\boldsymbol{h}_k,\boldsymbol{x})^{1-\alpha_n} \quad (\text{finite sum})$$
$$= \log\frac{1}{K}\sum_k q(\boldsymbol{h}_k|\boldsymbol{x})\left(\frac{p(\boldsymbol{h}_k,\boldsymbol{x})}{q(\boldsymbol{h}_k|\boldsymbol{x})}\right)^{1-\lim_{n\to\infty}\alpha_n} \quad (a^x \text{ is continuous in } x \text{ for all } a > 0)$$
$$= \log\frac{1}{K}\sum_k q(\boldsymbol{h}_k|\boldsymbol{x})^{\alpha} p(\boldsymbol{h}_k,\boldsymbol{x})^{1-\alpha}.$$

We note that since we assume $\hat{\mathcal{L}}_{\alpha,K}$ is absolutely integrable, we have $p/q > 0$ almost everywhere on the support of $q$. Hence $\{\hat{\mathcal{L}}_{\alpha_n,K}\}$ has point-wise limit $\hat{\mathcal{L}}_{\alpha,K}$ almost everywhere as $n \to +\infty$.

For $\alpha = 0, 1, \infty$ the Rényi divergence is defined by continuity so one can use the same technique to show the continuity of $\hat{\mathcal{L}}_{\alpha,K}$ on those $\alpha$ values for fixed $K$. Then since $\alpha_n \to \alpha$, for any $\epsilon > 0$, there

4

exists $n$ that is large enough such that $\alpha_m \in (\alpha - \epsilon, \alpha + \epsilon)$ for all $m > n$. Using the monotonicity result, we have for $\forall m > n$, $\hat{\mathcal{L}}_{\alpha_m,K}$ is bounded in the interval $(\hat{\mathcal{L}}_{\alpha+\epsilon,K}, \hat{\mathcal{L}}_{\alpha-\epsilon,K})$ and by assumption we have $\mathbb{E}[|\hat{\mathcal{L}}_{\alpha-\epsilon,K}|] < +\infty$ and $\mathbb{E}[|\hat{\mathcal{L}}_{\alpha+\epsilon,K}|] < +\infty$. This allows us to apply the dominated convergence theorem to prove $\lim_{n \to +\infty} \mathbb{E}[\hat{\mathcal{L}}_{\alpha_n,K}] = \mathbb{E}[\lim_{n \to +\infty} \hat{\mathcal{L}}_{\alpha_n,K}] = \mathbb{E}[\hat{\mathcal{L}}_{\alpha,K}]$. Thus we have proved that $\mathbb{E}[\hat{\mathcal{L}}_{\alpha,K}]$ is continuous on $\alpha \in \{|\mathcal{L}_\alpha| < +\infty\}$ if $\hat{\mathcal{L}}_{\alpha,K}$ is absolutely integrable. $\square$

### C.2 Proof of Corollary 1

It is sufficient to prove the corollary for the case $q(\boldsymbol{h}|\boldsymbol{x}) \neq p(\boldsymbol{h}|\boldsymbol{x})$. We first introduce the following lemmas. With overloaded notation, $\mu$ denotes the measure on the corresponding space, which also means $dQ = qd\mu$. As we assume $\mathrm{supp}(p) \subseteq \mathrm{supp}(q)$, there might exist some regions that $q > 0$ but $p = 0$. We define $\rho = \frac{\mu(\mathrm{supp}(q) \setminus \mathrm{supp}(p))}{\mu(\mathrm{supp}(q))}$ and rewrite the computation of $\mathbb{E}[\hat{\mathcal{L}}_{\alpha,K}]$.

**Lemma 1.** *Assume $\rho > 0$. Then for all finite $K$ and $\alpha < 0$, $\mathbb{E}_{\{\boldsymbol{h}_k\}_{k=1}^K}[\hat{\mathcal{L}}_{\alpha,K}(q;\boldsymbol{x})] = -\infty$ and thus $\hat{\mathcal{L}}_{\alpha,K}$ is not integrable wrt. $qd\mu = dQ$.*

*Proof.* We define $\tilde{q}$ as the $q$ distribution restricted on the support of $p$, i.e. $\tilde{q} = q/(1-\rho)$ defined on $\mathrm{supp}(p)$. Then for any fixed $K < +\infty$ and $\alpha < 0$, we have

$$\mathbb{E}_{\{\boldsymbol{h}_k\}_{k=1}^K \sim q}[\hat{\mathcal{L}}_{\alpha,K}(q;\boldsymbol{x})] = \rho^K \log 0 + \sum_{k=1}^K \binom{K}{k} \rho^{K-k}(1-\rho)^k \left(\mathbb{E}_{\{\boldsymbol{h}_j\}_{j=1}^k \sim \tilde{q}}[\hat{\mathcal{L}}_{\alpha,k}(\tilde{q};\boldsymbol{x})] + \log k\right)$$
$$- (1-\rho^K)((1-\alpha)\log(1-\rho) + \log K)$$

Thus $\mathbb{E}_{\{\boldsymbol{h}_k\}_{k=1}^K}[\hat{\mathcal{L}}_{\alpha,K}(q;\boldsymbol{x})] = -\infty$ for all finite $K$ and $\alpha < 0$. $\square$

The above example shows the pathology of MC approximation which is further discussed in section D. From now on we assume $\hat{\mathcal{L}}_{\alpha,K}$ is absolutely integrable in order to apply Theorem 2.

**Lemma 2.** *Assume $\alpha < 0$, $\hat{\mathcal{L}}_{\alpha,K}$ absolutely integrable wrt. $qd\mu = dQ$ for all $K$, $\mathcal{L}_\alpha > \mathcal{L}_{VI}$, and $|\mathcal{L}_\alpha| < +\infty$. Then there exists $1 \leq K_\alpha < +\infty$ such that for all $K \leq K_\alpha < K'$, $\mathbb{E}_{\{\boldsymbol{h}_k\}_{k=1}^K}[\hat{\mathcal{L}}_{\alpha,K}(q;\boldsymbol{x})] \leq \log p(\boldsymbol{x}) < \mathbb{E}_{\{\boldsymbol{h}_k\}_{k=1}^{K'}}[\hat{\mathcal{L}}_{\alpha,K'}(q;\boldsymbol{x})]$. Also $K_\alpha$ is **non-decreasing** in $\alpha$ with $\lim_{\alpha \to 0} K_\alpha = +\infty$ and $\lim_{\alpha \to -\infty} K_\alpha \geq 1$.*

*Proof.* 1) Existence of $K_\alpha$: first from Theorem 2 we have $\mathbb{E}[\hat{\mathcal{L}}_{\alpha,K}]$ is non-decreasing in $K$ when $\alpha < 0$. Then since for all $\alpha$, $\mathbb{E}[\hat{\mathcal{L}}_{\alpha,1}] = \mathcal{L}_{VI} \leq \log p(\boldsymbol{x})$, we have $K_\alpha \geq 1$ if $K_\alpha$ exists. Also from Theorem 2 we have $\lim_{K \to +\infty} \mathbb{E}[\hat{\mathcal{L}}_{\alpha,K}] = \mathcal{L}_\alpha > \log p(\boldsymbol{x})$ for all $\alpha < 0$. Hence for $\epsilon = \mathcal{L}_\alpha - \log p(\boldsymbol{x})$ there exist $K$ that is finite but large enough such that $\mathcal{L}_\alpha - \mathbb{E}[\hat{\mathcal{L}}_{\alpha,K'}] < \epsilon$ for all $K' > K$. Now we can define $\epsilon = \mathcal{L}_\alpha - \mathcal{L}_{VI}$ and take $K_\alpha$ as the minimum of such $K$, and it is straight-forward to show that $1 \leq K_\alpha < +\infty$.

2) $K_\alpha$ is non-decreasing in $\alpha$: suppose there exist $\alpha > \beta$ such that $K_\alpha < K_\beta$. Then there exist $K_\alpha < K \leq K_\beta$ such that $\mathbb{E}[\hat{\mathcal{L}}_{\alpha,K}] > \log p(\boldsymbol{x}) \geq \mathbb{E}[\hat{\mathcal{L}}_{\beta,K}]$. But Theorem 2 says $\mathbb{E}[\hat{\mathcal{L}}_{\alpha,K}]$ is non-increasing in $\alpha$, a contradiction.

3) Since $\lim_{K \to +\infty} \mathbb{E}[\hat{\mathcal{L}}_{\alpha,K}] = \mathcal{L}_\alpha$ and $\mathcal{L}_\alpha \downarrow \log p(\boldsymbol{x})$ when $\alpha \uparrow 0$, we have $\lim_{\alpha \to 0} K_\alpha = +\infty$. Also since $K_\alpha$ is non-decreasing in $\alpha$ and is lower-bounded by 1, we have the limit exists and $\lim_{\alpha \to -\infty} K_\alpha \geq 1$. $\square$

Now we prove Corollary 1, and we only prove it with the conditions assumed in Lemma 2 since $K_\alpha = +\infty$ for the other cases, and if so for all $\alpha < 0$, then $\alpha_K = -\infty$ for all finite $K$.

*Proof.* 1) Existence of $\alpha_K$ for $\lim_{\alpha \to -\infty} K_\alpha < K < +\infty$: from Lemma 2 we can find $\alpha > \beta$ such that $K_\alpha \geq K \geq K_\beta$. This means $\mathbb{E}[\hat{\mathcal{L}}_{\alpha,K}] \leq \log p(\boldsymbol{x}) \leq \mathbb{E}[\hat{\mathcal{L}}_{\beta,K}]$. Since $\mathbb{E}[\hat{\mathcal{L}}_{\alpha,K}]$ is continuous in $\alpha$ for any fixed $K$, there exits $\alpha \leq \gamma \leq \beta$ to have $\mathbb{E}[\hat{\mathcal{L}}_{\gamma,K}] = \log p(\boldsymbol{x})$. Note that $\gamma$ might not be

5

unique, so we define $\alpha_K$ as the minimum of such $\gamma$, which also gives $\mathbb{E}[\hat{\mathcal{L}}_{\alpha,K}] > \log p(\boldsymbol{x})$ for all $\alpha < \alpha_K$.

2) $\alpha_K$ is non-decreasing in $K$: suppose there exist $K < K'$ with $\alpha_K > \alpha_{K'}$. Then we can find $\alpha_K > \alpha > \alpha_{K'}$ such that $\mathbb{E}[\hat{\mathcal{L}}_{\alpha,K}] > \log p(\boldsymbol{x}) = \mathbb{E}[\hat{\mathcal{L}}_{\alpha_{K'},K'}] \geq \mathbb{E}[\hat{\mathcal{L}}_{\alpha,K'}]$. But from Theorem 2 $\mathbb{E}[\hat{\mathcal{L}}_{\alpha,K}]$ is non-decreasing in $K$, a contradiction.

3) Since $\lim_{K \to +\infty} \mathbb{E}[\hat{\mathcal{L}}_{\alpha,K}] = \mathcal{L}_\alpha$ and $\mathcal{L}_\alpha \downarrow \log p(\boldsymbol{x})$ when $\alpha \uparrow 0$, we have $\lim_{K \to +\infty} \alpha_K = 0$. Also for all $\alpha$, $\mathbb{E}[\hat{\mathcal{L}}_{\alpha,1}] = \mathcal{L}_{VI} \leq \log p(\boldsymbol{x})$, so $\lim_{K \to 1} \alpha_K = -\infty$.

□

## D  Optimisation issues with $\alpha$-divergences and MC approximations

It is in general an outstanding research question on how to select the divergence measure for a particular machine learning problem. In our case this corresponds to selecting the $\alpha$ value. Also an approximate inference algorithm can be evaluated with different performance measures, and it is impossible to find a single algorithm value that returns the best performance on all evaluations. Thus we only present the evaluation in test error and test log-likelihood in the main text.

We discuss two conjectures to explain the difficulty of selecting $\alpha$ in the Bayesian neural network experiments. The first conjecture is that zero-forcing algorithms tend to favour minimising the test error, while mass-covering methods tend to improve the test log-likelihood. However zero-forcing methods can fail as it might miss an important mode due to local optima. Similarly mass-covering methods can be pathological if the exact posterior includes modes that are very far away from each other. Furthermore, the form of the posterior will change with the number of observed datapoints $N$, so the "optimal" setting of $\alpha$ for a fixed task may change with $N$.

The second conjecture states that the MC approximation complicates the selection of $\alpha$, since it favours zero-forcing (because of the bias introduced). For example, in order to maximize the quantity of the MC approximation the algorithm need to make $\mathbb{E}[\hat{\mathcal{L}}_{\alpha,K}]$ finite first. However, Lemma 1 indicates that, if $\rho > 0$, then for finite sample size, there's a small probability $\rho^K$ that the MC approximation goes wrong. Hence to avoid this pathology the optimisation procedure will ensure $q = 0$ whenever $p$ is zero. Combining with Theorem 2, we conjecture that the MC approximation makes the algorithm more "VI-like" compared to the exact case. In other words, when MC approximation is deployed, the effective $\alpha$ value is closer to $\alpha = 1$ which is the value for VI (consider $K = 1$). This means, if there exists $\alpha_{\text{opt}} \neq 1$ for a specific task, in practice one should use $\alpha \leq \alpha_{\text{opt}}$ (for $\alpha_{\text{opt}} < 1$, and should use $\alpha \geq \alpha_{\text{opt}}$ if $\alpha_{\text{opt}} > 1$) when running the MC algorithm. In general one should be very careful when estimating the ratio between distribution with Monte Carlo methods. Also the introduced MC approach usually has higher variance compared to the variational case, so further control variate techniques should be applied to reduce the sampling variance.

Still we want to emphasize that for many problems, minimising an $\alpha$-divergence other than the KL-divergence can be very useful, even when with MC approximations. Approximate EP has been applied to deep Gaussian process regression and has shown to achieve the state-of-the-art results for benchmark datasets [7]. A recent paper [8] tested BB-$\alpha$ for model-based reinforcement learning with Bayesian neural networks. In their tests using $\alpha = 0.5$ successfully captured the bi-modality and heteroskedasticity in the predictive distribution, while VI failed disastrously.

## E  Unified implementation: derivation details

We provide detailed derivations of the gradient computation here. Recall from the main text that when $\alpha \neq 1$, the VR bound with the reparameterization trick becomes

$$\mathcal{L}_\alpha(q_\phi; \boldsymbol{x}) = \frac{1}{1-\alpha} \log \mathbb{E}_{\boldsymbol{\epsilon}} \left[ \left( \frac{p(g_\phi(\boldsymbol{\epsilon}), \boldsymbol{x})}{q(g_\phi(\boldsymbol{\epsilon}))} \right)^{1-\alpha} \right]. \tag{2}$$

So the distribution $p(\epsilon)$ does not depend on the recognition model. We short-hand $g_\phi = g_\phi(\epsilon)$, then,

$$\nabla_\phi \mathcal{L}_\alpha(q_\phi; \boldsymbol{x}) = \frac{1}{1-\alpha} \nabla_\phi \log \mathbb{E}_\epsilon \left[ \left( \frac{p(g_\phi, \boldsymbol{x})}{q(g_\phi)} \right)^{1-\alpha} \right]$$

$$= \frac{1}{1-\alpha} \left( \mathbb{E}_\epsilon \left[ \left( \frac{p(g_\phi, \boldsymbol{x})}{q(g_\phi)} \right)^{1-\alpha} \right] \right)^{-1} \mathbb{E}_\epsilon \left[ \nabla_\phi \left( \frac{p(g_\phi, \boldsymbol{x})}{q(g_\phi)} \right)^{1-\alpha} \right]$$

$$= \frac{1}{1-\alpha} \left( \mathbb{E}_\epsilon \left[ \left( \frac{p(g_\phi, \boldsymbol{x})}{q(g_\phi)} \right)^{1-\alpha} \right] \right)^{-1} \mathbb{E}_\epsilon \left[ \left( \frac{p(g_\phi, \boldsymbol{x})}{q(g_\phi)} \right)^{1-\alpha} \nabla_\phi (1-\alpha) \log \frac{p(g_\phi, \boldsymbol{x})}{q(g_\phi)} \right]$$

$$= \mathbb{E}_\epsilon \left[ w_\alpha(\epsilon; \phi, \boldsymbol{x}) \nabla_\phi \log \frac{p(g_\phi, \boldsymbol{x})}{q(g_\phi)} \right].$$

Here we define

$$w_\alpha(\epsilon; \phi, \boldsymbol{x}) := \left( \frac{p(g_\phi, \boldsymbol{x})}{q(g_\phi)} \right)^{1-\alpha} \Big/ \mathbb{E}_\epsilon \left[ \left( \frac{p(g_\phi, \boldsymbol{x})}{q(g_\phi)} \right)^{1-\alpha} \right]. \quad (3)$$

For MC approximation with finite $K$ samples, one can use the same technique to show that

$$\nabla_\phi \hat{\mathcal{L}}_{\alpha, K}(q_\phi; \boldsymbol{x}) = \sum_{k=1}^{K} \left[ \hat{w}_{\alpha, k}(\epsilon_k; \phi, \boldsymbol{x}) \nabla_\phi \log \frac{p(g_\phi(\epsilon_k), \boldsymbol{x})}{q(g_\phi(\epsilon_k))} \right].$$

with the importance weights

$$\hat{w}_{\alpha, k}(\epsilon_k; \phi, \boldsymbol{x}) := \left( \frac{p(g_\phi(\epsilon_k), \boldsymbol{x})}{q(g_\phi(\epsilon_k))} \right)^{1-\alpha} \Big/ \sum_{k=1}^{K} \left( \frac{p(g_\phi(\epsilon_k), \boldsymbol{x})}{q(g_\phi(\epsilon_k))} \right)^{1-\alpha}. \quad (4)$$

One can show that $\lim_{\alpha \to 1} w_\alpha(\epsilon; \phi, \boldsymbol{x}) = 1$ and $\lim_{\alpha \to 1} \hat{w}_{\alpha, k}(\epsilon_k; \phi, \boldsymbol{x}) = 1/K$. This indicates the recovery of the original VAE algorithm.

## F   Stochastic approximation for large-scale learning: derivations

This section shows the connection between VR bound optimisation and the recently proposed algorithms: SEP [1] and BB-$\alpha$ [2], by taking $M = 1$ and $\alpha = 1 - \beta/N$.

Recall that in the main text we define the "average likelihood" $\bar{f}_\mathcal{D}(\boldsymbol{\theta}) = [\prod_{n=1}^{N} p(\boldsymbol{x}_n | \boldsymbol{\theta})]^{\frac{1}{N}}$. Hence the joint distribution can be rewritten as $p(\boldsymbol{\theta}, \mathcal{D}) = p_0(\boldsymbol{\theta}) \bar{f}_\mathcal{D}(\boldsymbol{\theta})^N$. Also for a minibatch of $M$ datapoints $\mathcal{S} = \{\boldsymbol{x}_{n_1}, ..., \boldsymbol{x}_{n_m}\} \sim \mathcal{D}$, we define the "subset average likelihood" $\bar{f}_\mathcal{S} = [\prod_{m=1}^{M} p(\boldsymbol{x}_{n_m} | \boldsymbol{\theta})]^{\frac{1}{M}}$. When $M = 1$ we also write $\bar{f}_\mathcal{S}(\boldsymbol{\theta}) = f_n(\boldsymbol{\theta})$ for $\mathcal{S} = \{\boldsymbol{x}_n\}$.

Now assume the posterior approximation is defined as $q(\boldsymbol{\theta}) = \frac{1}{Z_q} p_0(\boldsymbol{\theta}) t(\boldsymbol{\theta})^N$. Often $t(\boldsymbol{\theta})$ is chose to have an exponential family form $t(\boldsymbol{\theta}) \propto \exp[\langle \boldsymbol{\lambda}, \boldsymbol{\Phi}(\boldsymbol{\theta}) \rangle]$ with $\boldsymbol{\Phi}(\boldsymbol{\theta})$ denoting the sufficient statistic. Then picking $\alpha = 1 - \beta/N, \beta \neq 0$, we have the exact VR bound as

$$\mathcal{L}_\alpha(q; \mathcal{D}) = \log Z_q + \frac{N}{\beta} \log \mathbb{E}_q \left[ \left( \frac{\bar{f}_\mathcal{D}(\boldsymbol{\theta})}{t(\boldsymbol{\theta})} \right)^\beta \right] \quad (5)$$

The first proposal considers deriving the exact fixed point conditions, then approximating them with mini-batch sub-sampling. In our example the exact fixed point condition for the variational parameters $\boldsymbol{\lambda}$ is

$$\nabla_{\boldsymbol{\lambda}} \mathcal{L}_\alpha(q; \mathcal{D}) = 0 \quad \Rightarrow \quad \mathbb{E}_q[\boldsymbol{\Phi}(\boldsymbol{\theta})] = \mathbb{E}_{\tilde{p}_\alpha}[\boldsymbol{\Phi}(\boldsymbol{\theta})], \quad (6)$$

with the tilted distribution defined as

$$\tilde{p}_\alpha(\boldsymbol{\theta}) \propto q(\boldsymbol{\theta})^\alpha p_0(\boldsymbol{\theta})^{1-\alpha} \bar{f}_\mathcal{D}(\boldsymbol{\theta})^{N(1-\alpha)} \propto p_0(\boldsymbol{\theta}) t(\boldsymbol{\theta})^{N-\beta} \bar{f}_\mathcal{D}(\boldsymbol{\theta})^\beta.$$

Now given a mini-batch of datapoints $\mathcal{S}$, the moment matching update can be approximated by replacing $\bar{f}_\mathcal{D}(\boldsymbol{\theta})$ with $\bar{f}_\mathcal{S}(\boldsymbol{\theta}) = [\prod_{m=1}^{M} p(\boldsymbol{x}_{n_m} | \boldsymbol{\theta})]^{\frac{1}{M}}$. More precisely, each iteration we sample a

subset of data $\mathcal{S} = \{\boldsymbol{x}_{n_1}, ..., \boldsymbol{x}_{n_M}\} \sim \mathcal{D}$, and compute the new update for $\boldsymbol{\lambda}$ by first computing $\tilde{p}_{\alpha, \mathcal{S}}(\boldsymbol{\theta}) \propto p_0(\boldsymbol{\theta}) t(\boldsymbol{\theta})^{N-\beta} \bar{f}_{\mathcal{S}}(\boldsymbol{\theta})^{\beta}$ then taking $\mathbb{E}_q[\boldsymbol{\Phi}(\boldsymbol{\theta})] \leftarrow \mathbb{E}_{\tilde{p}_{\alpha, \mathcal{S}}}[\boldsymbol{\Phi}(\boldsymbol{\theta})]$. This method returns SEP when $M = 1$, i.e. in each iteration only one datapoint is sampled to update the approximate posterior.

The second proposal also applies this subset average likelihood approximation idea, but directly to the VR bound (5), with $\mathbb{E}_{\mathcal{S}}$ denotes the expectation over mini-batch sub-sampling:

$$\mathbb{E}_{\mathcal{S}}\left[\tilde{\mathcal{L}}_\alpha(q; \mathcal{S})\right] = \log Z_q + \frac{N}{\beta} \mathbb{E}_{\mathcal{S}} \left[\log \mathbb{E}_q \left[\left(\frac{\bar{f}_{\mathcal{S}}(\boldsymbol{\theta})}{t(\boldsymbol{\theta})}\right)^\beta\right]\right]. \quad (7)$$

It recovers the energy function of BB-$\alpha$ when $M = 1$. Note that the original paper [2] uses an adapted form of Amari's $\alpha$-divergence, and the $\alpha$ value in the BB-$\alpha$ algorithm corresponds to $\beta$ in our exposition. Now the gradient of this approximated energy function becomes

$$\nabla_{\boldsymbol{\lambda}} \mathbb{E}_{\mathcal{S}}\left[\tilde{\mathcal{L}}_\alpha(q; \mathcal{S})\right] = N(\mathbb{E}_q[\boldsymbol{\Phi}(\boldsymbol{\theta})] - \mathbb{E}_{\mathcal{S}} \mathbb{E}_{\tilde{p}_{\alpha, \mathcal{S}}}[\boldsymbol{\Phi}(\boldsymbol{\theta})]). \quad (8)$$

Both SEP and BB-$\alpha$ return SVI when $\alpha \to 1$ (or equivalently $\beta \to 0$). But for other $\alpha$ values it is important to note that these two proposals return different optimum at convergence. BB-$\alpha$ requires averages the moment of the tilted distribution $\mathbb{E}_{\mathcal{S}} \mathbb{E}_{\tilde{p}_{\alpha, \mathcal{S}}}[\boldsymbol{\Phi}(\boldsymbol{\theta})]$. However SEP first compute the inverse mapping from the moment $\mathbb{E}_{\tilde{p}_{\alpha, \mathcal{S}}}[\boldsymbol{\Phi}(\boldsymbol{\theta})]$ to obtain the natural parameters $\boldsymbol{\lambda}_{\mathcal{S}}$, then update the $q$ distribution by $\boldsymbol{\lambda} \leftarrow \mathbb{E}_{\mathcal{S}}[\boldsymbol{\lambda}_{\mathcal{S}}]$. In general the inverse mapping is non-linear so the fixed point conditions of SEP and BB-$\alpha$ are different.

SEP is arguably more well justified since it returns the exact posterior if the approximation family $\mathcal{Q}$ is large enough to include the correct solution, just like VI and VR computed on the whole dataset. BB-$\alpha$ might still be biased even in this scenario. But BB-$\alpha$ is much simpler to implement since the energy function can be optimised with stochastic gradient descent. Indeed the authors of [2] considered the same black-box approach as to VI, by computing a stochastic estimate of the energy function then using automatic differentiation tools to obtain the gradients.

We also provide a bound of the energy approximation (7) by the following theorem.

**Theorem 3.** *If the approximate distribution $q(\boldsymbol{\theta})$ is Gaussian $\mathcal{N}(\boldsymbol{\mu}, \boldsymbol{\Sigma})$, and the likelihood functions has an exponential family form $p(\boldsymbol{x}|\boldsymbol{\theta}) = \exp[\langle \boldsymbol{\theta}, \boldsymbol{\Psi}(\boldsymbol{x}) \rangle - A(\boldsymbol{\theta})]$, then for $\alpha \leq 1$ and $r > 1$ the stochastic approximation is bounded by*

$$\mathbb{E}_{\mathcal{S}}[\tilde{\mathcal{L}}_\alpha(q; \mathcal{S})] \leq \mathcal{L}_{1-(1-\alpha)r}(q; \mathcal{D}) + \frac{N^2(1-\alpha)r}{2(r-1)} \text{tr}(\boldsymbol{\Sigma} \text{Cov}_{\mathcal{S} \sim \mathcal{D}}(\bar{\boldsymbol{\Psi}}_{\mathcal{S}})).$$

*Proof.* We substitute the exponential family likelihood term into the stochastic approximation of the VR bound with $\alpha < 1$, and use Hölder's inequality for any $1/r + 1/s = 1$, $r > 1$ (define $\tilde{\alpha} = 1 - (1-\alpha)r$):

$$\mathbb{E}_{\mathcal{S}}[\tilde{\mathcal{L}}_\alpha(q; \mathcal{S})] = \frac{1}{1-\alpha} \log \mathbb{E}_q\left[\left(\frac{p_0(\boldsymbol{\theta}) \bar{f}_{\mathcal{D}}(\boldsymbol{\theta})^N}{q(\boldsymbol{\theta})} \frac{\bar{f}_{\mathcal{S}}(\boldsymbol{\theta})^N}{\bar{f}_{\mathcal{D}}(\boldsymbol{\theta})^N}\right)^{1-\alpha}\right]$$

$$\leq \mathcal{L}_{\tilde{\alpha}}(q; \mathcal{D}) + \frac{1}{(1-\alpha)s} \mathbb{E}_{\mathcal{S}} \left\{\log \mathbb{E}_q[\exp[N(1-\alpha)s \langle \bar{\boldsymbol{\Psi}}_{\mathcal{S}} - \bar{\boldsymbol{\Psi}}_{\mathcal{D}}, \boldsymbol{\theta}\rangle]]\right\}$$

$$= \mathcal{L}_{\tilde{\alpha}}(q; \mathcal{D}) + \frac{1}{(1-\alpha)s} \mathbb{E}_{\mathcal{S}}[K_{\boldsymbol{\theta}}(N(1-\alpha)s(\bar{\boldsymbol{\Psi}}_{\mathcal{S}} - \bar{\boldsymbol{\Psi}}_{\mathcal{D}}))],$$

where $\bar{\boldsymbol{\Psi}}_{\mathcal{S}}$ and $\bar{\boldsymbol{\Psi}}_{\mathcal{D}}$ denote the mean of the sufficient statistic $\boldsymbol{\Psi}(\boldsymbol{x})$ on the mini-batch $\mathcal{S}$ and the whole dataset $\mathcal{D}$, respectively. For Gaussian distribution $q(\boldsymbol{\theta}) = \mathcal{N}(\boldsymbol{\mu}, \boldsymbol{\Sigma})$ the cumulant generating function $K_{\boldsymbol{\theta}}(\boldsymbol{t})$ has a closed form

$$K_{\boldsymbol{\theta}}(\boldsymbol{t}) = \boldsymbol{\mu}^T \boldsymbol{t} + \frac{1}{2} \boldsymbol{t}^T \boldsymbol{\Sigma} \boldsymbol{t}.$$

Define $t_\mathcal{S} = N(1-\alpha)s\Delta_\mathcal{S}$ with $\Delta_\mathcal{S} = \bar{\Psi}_\mathcal{S} - \bar{\Psi}_\mathcal{D}$, then $\mathbb{E}_\mathcal{S}[t_\mathcal{S}] = \mathbf{0}$ and the upper-bound becomes

$$\mathbb{E}_\mathcal{S}[\tilde{\mathcal{L}}_\alpha(q;\mathcal{S})] \leq \mathcal{L}_{\tilde{\alpha}}(q;\mathcal{D}) + \frac{1}{(1-\alpha)s}\mathbb{E}_\mathcal{S}[K_\theta(t_\mathcal{S})]$$

$$= \mathcal{L}_{\tilde{\alpha}}(q;\mathcal{D}) + \frac{1}{(1-\alpha)s}\mathbb{E}_\mathcal{S}[\boldsymbol{\mu}^T t_\mathcal{S} + \frac{1}{2}t_\mathcal{S}^T \boldsymbol{\Sigma} t_\mathcal{S}]$$

$$= \mathcal{L}_{\tilde{\alpha}}(q;\mathcal{D}) + \frac{N^2(1-\alpha)s}{2}\mathbb{E}_\mathcal{S}[\Delta_\mathcal{S}^T \boldsymbol{\Sigma} \Delta_\mathcal{S}]$$

$$= \mathcal{L}_{\tilde{\alpha}}(q;\mathcal{D}) + \frac{N^2(1-\alpha)s}{2}\text{tr}(\boldsymbol{\Sigma}\text{Cov}_{\mathcal{S}\sim\mathcal{D}}(\bar{\boldsymbol{\Psi}}_\mathcal{S})).$$

Applying the condition of Hölder's inequality $1/r + 1/s = 1$ proves the result. □

The following corollary is a direct result of Theorem 3 applied to BB-$\alpha$. Note here we follow the convention of the original paper [2] to use $M = 1$ and overload the notation $\alpha = \beta$ and $\mathcal{L}_{BB-\alpha}(q;\mathcal{D}) = \mathbb{E}_{\{\boldsymbol{x}_n\}}\left[\tilde{\mathcal{L}}_{1-\alpha/N}(q;\{\boldsymbol{x}_n\})\right]$.

**Corollary 2.** *Assume the approximate posterior and the likelihood functions satisfy the assumptions in Theorem 3, then for $\alpha > 0$ and $r > 1$, the black-box alpha energy function is upper-bounded by*

$$\mathcal{L}_{BB-\alpha}(q;\mathcal{D}) \leq \mathcal{L}_{1-\frac{\alpha r}{N}}(q;\mathcal{D}) + \frac{N\alpha r}{2(r-1)}\text{tr}(\boldsymbol{\Sigma}\text{Cov}_\mathcal{D}(\boldsymbol{\Psi})).$$

## G  Further experimental details and results

### G.1  Bayesian neural network

We detail the experimental set-up of the Bayesian neural network example. For regression tests, we consider Protein and Year as the large datasets and the others as small datasets. The likelihood function is defined as $p(y|\boldsymbol{x},\boldsymbol{\theta}) = \mathcal{N}(y; F_{\boldsymbol{\theta}}(\boldsymbol{x}), \sigma^2)$ where $F_{\boldsymbol{\theta}}(\boldsymbol{x})$ denotes the non-linear transform from the neural network with weights $\boldsymbol{\theta}$. We use unit Gaussian prior $\boldsymbol{\theta} \sim \mathcal{N}(\boldsymbol{\theta}; \mathbf{0}, \boldsymbol{I})$ and Gaussian approximation $q(\boldsymbol{\theta}) = \mathcal{N}(\boldsymbol{\theta}; \boldsymbol{\mu}_q, diag(\boldsymbol{\sigma}_q))$, where we fit the parameters of $q$ and the noise level $\sigma$ by optimising the lower-bound. For all datasets we use single-layer neural networks with 50 hidden units (ReLUs) for datasets except Protein and Year (100 units). The methods for comparison were run for 500 epochs on the small datasets and 100, 40 epochs for the large datasets Protein and Year, respectively. We used ADAM [9] for optimisation with learning rate 0.001 and the standard setting for other parameters. For stochastic optimisation we used learning rate 0.001, mini-batch size $M = 32$ and number of samples $K = 100, 10$ for small and large datasets. The number of dataset random splits is 20 except for the large datasets, which is 5 and 1 for Protein and Year, respectively.

The full test results are provided in Figure 1 and Table 1, 2. In the tables the best performing results are underlined, while the worse cases are also bold-faced. Clearly the optimal $\alpha$ setting is dataset dependent, although for Boston and Power the performances are very similar. Also for Naval mass-covering seems to be harmful not only for predictive error but also for test log-likelihood measure. Overall mode-seeking methods tend to focus on improving the predictive error, while mass-covering regimes often return better test log-likelihood.

### G.2  Variational auto-encoder

We describe the network architecture tested in the VAE experiments. The number of stochastic layers $L$, number of hidden units, and the activation function are summarised in Table 3. The prefix of the number indicates whether this layer is **d**eterministic or **s**tochastic, e.g. d500-s200 stands for a neural network with one deterministic layer of 500 units followed by a stochastic layer of 200 units. For Frey Face data we train the models using learning rate 0.0005 and mini-batch size 100. For MNIST and OMNIGLOT we reuse the settings from [5]: the training process runs for $3^i$ passes with learning rate $0.0001 \cdot 10^{-i/7}$ for $i = 0, ..., 7$, and the batch size is 20. For Caltech Silhouettes we use the same settings as MNIST and OMNIGLOT except that the training proceeded for $\sum_{i=0}^{7} 2^i = 255$ epochs.

We also present some samples from the VR-max trained auto-encoders in Figure 2, and note that the visual quality of these samples are almost identical to those from IWAE.

9

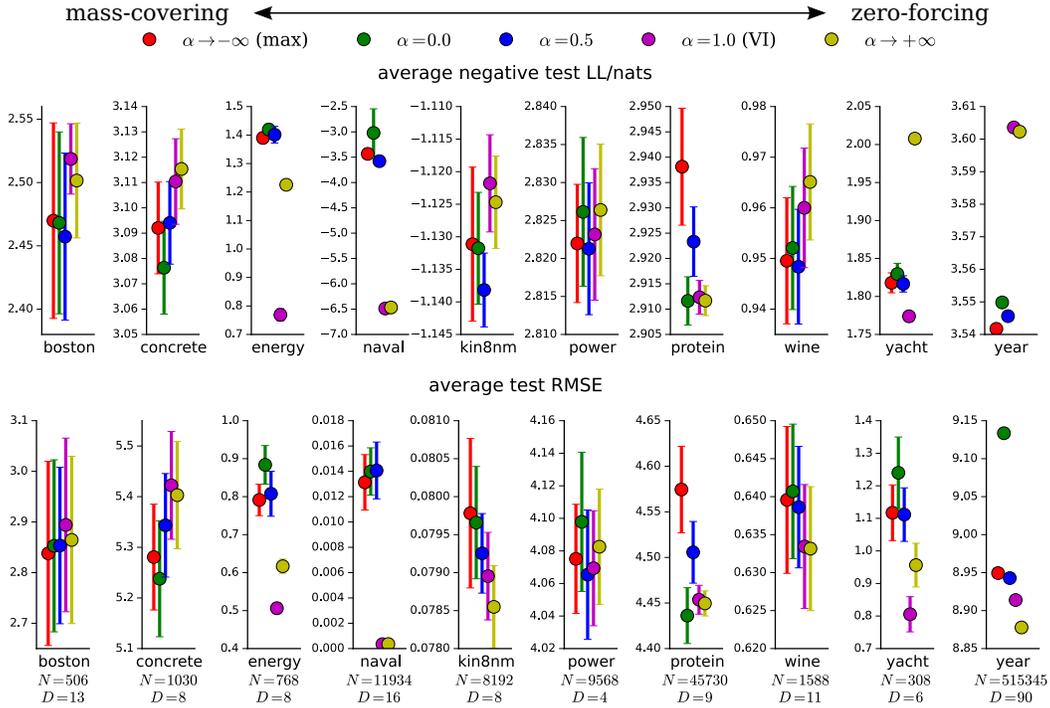

Figure 1: Test LL and RMSE results for Bayesian neural network regression. The lower the better.

Table 1: Regression experiment: Average negative test log likelihood/nats

| **Dataset** | N | D | $\alpha \to -\infty$ | $\alpha = 0.0$ | $\alpha = 0.5$ | $\alpha = 1.0$ **(VI)** | $\alpha \to +\infty$ |
|---|---|---|---|---|---|---|---|
| boston | 506 | 13 | 2.47±0.08 | 2.47±0.07 | *2.46±0.07* | **2.52±0.03** | 2.50±0.05 |
| concrete | 1030 | 8 | 3.09±0.02 | *3.08±0.02* | 3.09±0.02 | 3.11±0.02 | **3.12±0.02** |
| energy | 768 | 8 | 1.39±0.02 | **1.42±0.02** | 1.40±0.03 | *0.77±0.02* | 1.23±0.01 |
| naval | 11934 | 16 | -3.43±0.08 | **-3.02±0.48** | -3.58±0.08 | *-6.49±0.04* | -6.47±0.09 |
| kin8nm | 8192 | 8 | -1.13±0.01 | -1.13±0.01 | *-1.14±0.01* | **-1.12±0.01** | -1.12±0.01 |
| power | 9568 | 4 | 2.82±0.01 | 2.83±0.01 | *2.82±0.01* | 2.82±0.01 | **2.83±0.01** |
| protein | 45730 | 9 | **2.94±0.01** | *2.91±0.00* | 2.92±0.01 | 2.91±0.00 | 2.91±0.00 |
| wine | 1588 | 11 | 0.95±0.01 | 0.95±0.01 | *0.95±0.01* | 0.96±0.01 | **0.97±0.01** |
| yacht | 308 | 6 | 1.82±0.01 | 1.83±0.01 | 1.82±0.01 | *1.77±0.01* | **2.01±0.00** |
| year | 515345 | 90 | *3.54*±NA | 3.55±NA | 3.55±NA | **3.60**±NA | 3.60±NA |
| **Average Rank** | | | 2.80±0.34 | 3.00±0.45 | **2.20±0.37** | 3.20±0.51 | 3.80±0.39 |

Table 2: Regression experiment: Average test RMSE

| **Dataset** | N | D | $\alpha \to -\infty$ | $\alpha = 0.0$ | $\alpha = 0.5$ | $\alpha = 1.0$ **(VI)** | $\alpha \to +\infty$ |
|---|---|---|---|---|---|---|---|
| boston | 506 | 13 | *2.84±0.18* | 2.85±0.17 | 2.85±0.15 | **2.89±0.17** | 2.86±0.17 |
| concrete | 1030 | 8 | 5.28±0.10 | *5.24±0.11* | 5.34±0.10 | **5.42±0.11** | 5.40±0.11 |
| energy | 768 | 8 | 0.79±0.04 | **0.88±0.05** | 0.81±0.06 | *0.51±0.01* | 0.62±0.02 |
| naval | 11934 | 16 | 0.01±0.00 | 0.01±0.00 | **0.01±0.00** | *0.00±0.00* | 0.00±0.00 |
| kin8nm | 8192 | 8 | **0.08±0.00** | 0.08±0.00 | 0.08±0.00 | 0.08±0.00 | *0.08±0.00* |
| power | 9568 | 4 | 4.08±0.03 | **4.10±0.04** | *4.07±0.04* | 4.07±0.04 | 4.08±0.04 |
| protein | 45730 | 9 | **4.57±0.05** | *4.44±0.03* | 4.51±0.03 | 4.45±0.02 | 4.45±0.01 |
| wine | 1588 | 11 | 0.64±0.01 | **0.64±0.01** | 0.64±0.01 | 0.63±0.01 | *0.63±0.01* |
| yacht | 308 | 6 | 1.12±0.09 | **1.24±0.11** | 1.11±0.08 | *0.81±0.05* | 0.96±0.07 |
| year | 515345 | 90 | 8.95±NA | **9.13**±NA | 8.94±NA | 8.91±NA | *8.88*±NA |
| **Average Rank** | | | 3.40±0.38 | 3.70±0.51 | 3.20±0.31 | 2.40±0.45 | **2.30±0.38** |

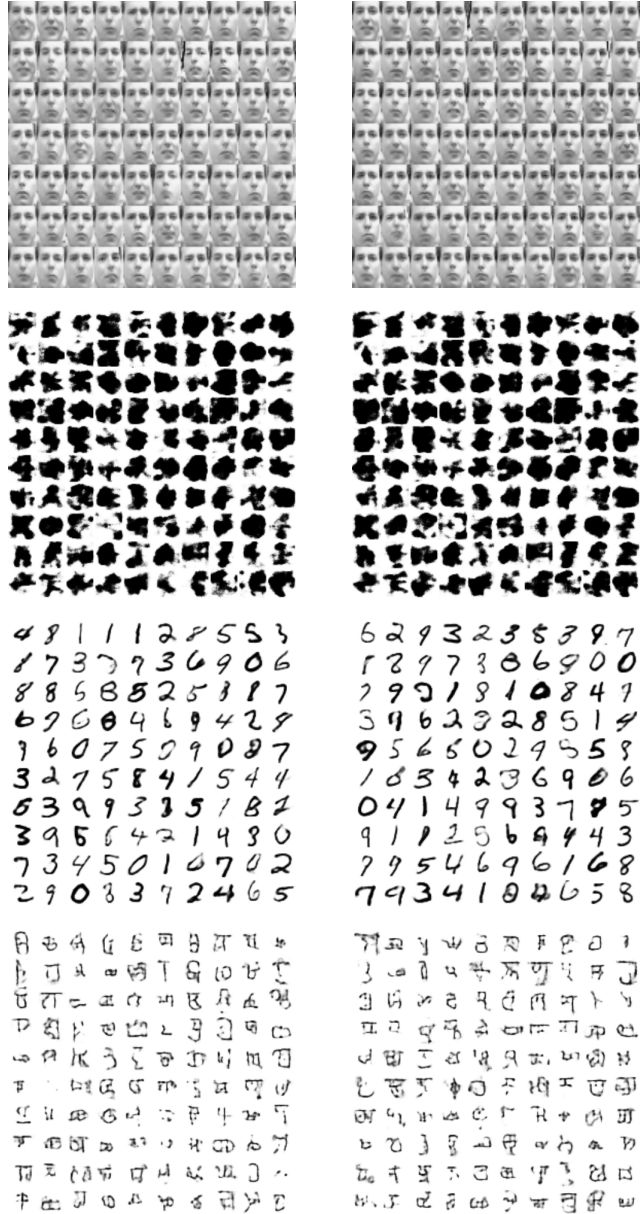

Figure 2: Sampled images from the the best models trained with IWAE (left) and VR-max (right).

Table 3: Network architecture of tested VAE algorithms.

| Dataset | L | architecture | activation | probability type (p/q) |
|---|---|---|---|---|
| Frey Face | 1 | d200-d200-s20 | softplus | Gaussian/Gaussian |
| Caltech 101 | 1 | d500-s200 | tanh | Bernoulli/Gaussian |
| MNIST & | 1 | d200-d200-s50 | tanh | Bernoulli/Gaussian |
| OMNIGLOT | 2 | d200-d200-s100-d100-d100-s50 | tanh | Bernoulli/Gaussian |



\end{document}